%% file: main.tex
\documentclass[10pt,twocolumn,letterpaper]{article}

\usepackage{cvpr}              

\usepackage{url}
\usepackage{xcolor} 
\usepackage{graphicx}
\usepackage{tabularx}
\usepackage{booktabs}
\usepackage{float}
\usepackage{caption}
\usepackage{colortbl}
\usepackage{wrapfig} 
\usepackage{multirow}
\usepackage{siunitx}
\usepackage{booktabs}
\usepackage{makecell}
\usepackage{dashrule}
\usepackage{subcaption}
\usepackage{algorithm}
\usepackage{algorithmic}
\usepackage[accsupp]{axessibility}  

\definecolor{cvprblue}{rgb}{0.21,0.49,0.74}
\definecolor{tabfirst}{RGB}{200, 225, 255}   
\definecolor{tabsecond}{RGB}{225, 240, 255}  
\definecolor{tabthird}{RGB}{242, 248, 255}   

\newcommand{\cellfirst}[1]{\cellcolor{tabfirst}{#1}}
\newcommand{\cellsecond}[1]{\cellcolor{tabsecond}{#1}}
\newcommand{\cellthird}[1]{\cellcolor{tabthird}{#1}}

\input{preamble}
\definecolor{cvprblue}{rgb}{0.21,0.49,0.74}
\usepackage[pagebackref,breaklinks,colorlinks,allcolors=cvprblue]{hyperref}

\def\paperID{} 
\def\confName{CVPR}
\def\confYear{2026}

\title{GLINT: Modeling Scene-Scale Transparency via Gaussian Radiance Transport}

\author{
Youngju Na$^{1,2,\ast}$\,\,\,\,\,
Jaeseong Yun$^{2}$\,\,\,\,\,
Soohyun Ryu$^{2}$\,\,\,\,\,
Hyunsu Kim$^{2}$\,\,\,\,\,
Sung-Eui Yoon$^{1}$\,\,\,\,\,
Suyong Yeon$^{2}$
\\\\
$^1$KAIST\,\,\,\,\,\,\,\,\,\,\,\,\,\,\,\,\,\,$^2$NAVER LABS
\\\\
\url{https://youngju-na.github.io/GLINT}
}

\begin{document}
\twocolumn[
    \maketitle
    \vspace{-2em}
    \input{figures/1_conceptual}

    \bigbreak
]
\renewcommand{\thefootnote}{\fnsymbol{footnote}}
\footnotetext[1]{Work done during an internship at NAVER LABS.}
\input{contents/0_abstract} 
\input{contents/1_intro}

\input{contents/2_related}

\input{contents/3_method}

\input{contents/4_experiments}
\input{contents/5_discussion}

\section{Acknowledgements}
We sincerely appreciate the reviewers for their thoughtful comments.
We also thank Kyu Beom Han and Taeyeon Kim for their careful proofreading.
This work was supported in part by the Institute of Information \& Communications Technology Planning \& Evaluation (IITP) grant funded by the Korea government (MSIT) (No. RS-2025-25443318, Physically-grounded Intelligence: A Dual Competency Approach to Embodied AGI through Constructing and Reasoning in the Real World), in part by the National Research Foundation of Korea (NRF) grant funded by the Korea government (MSIT) (No. RS-2023-00208506), and in part by the Korea Planning \& Evaluation Institute of Industrial Technology (KEIT) and the Ministry of Trade, Industry \& Resources (MOTIR) of the Republic of Korea (No. RS-2024-00417108).

{
    \small
    \bibliographystyle{ieeenat_fullname}
    \bibliography{main}
}

\input{contents/X_suppl}

\end{document}

%% file: figures/1_conceptual.tex
\begin{center}
    \captionsetup{type=figure}
    \vspace{-5pt}
    \includegraphics[width=1.0\textwidth, trim=0.1cm 4.5cm 4.373cm 0,clip]{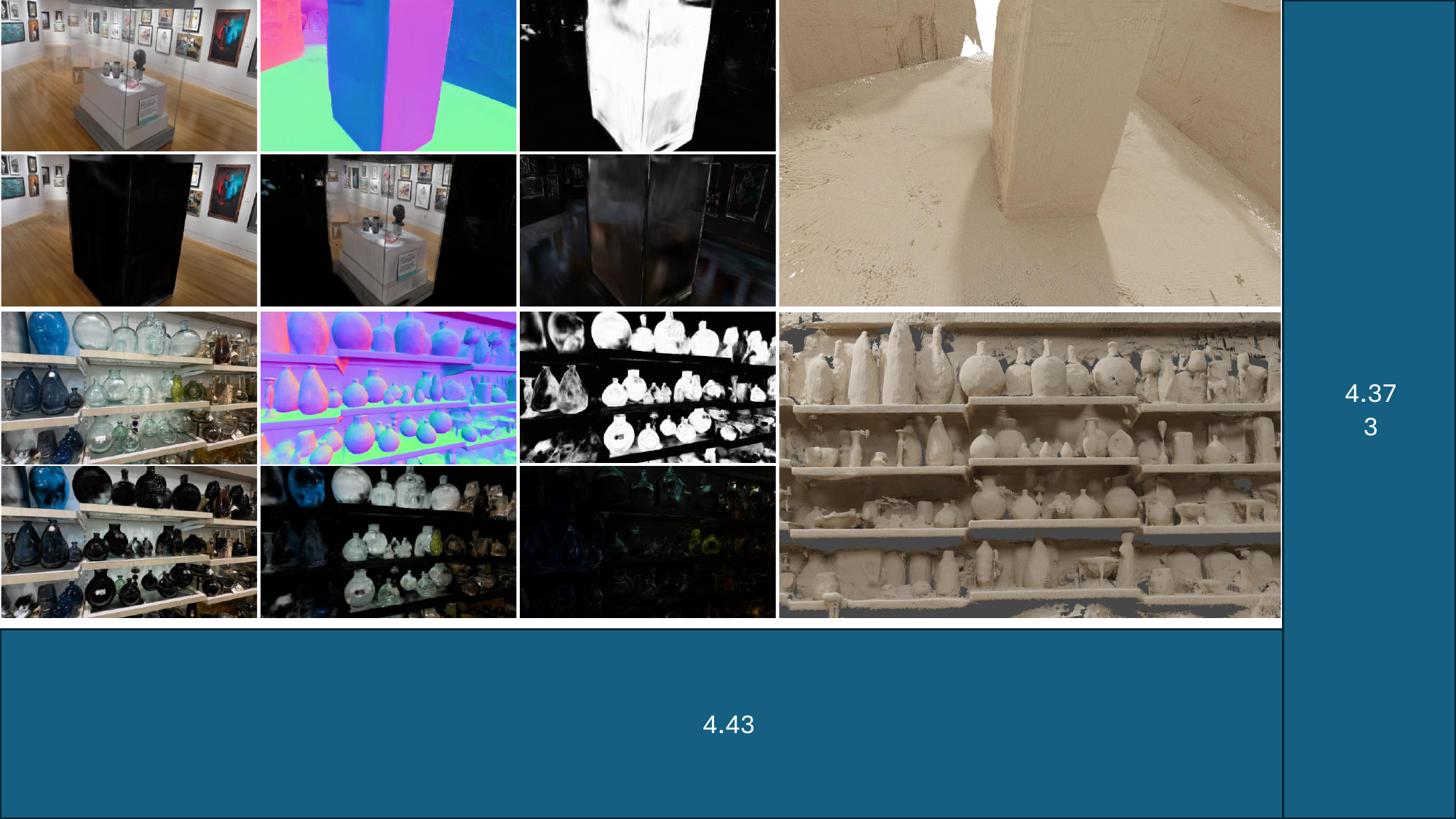}
    \vspace{-15pt}
    \caption{
    Our framework GLINT performs decomposed Gaussian radiance transport to reconstruct transparent surfaces with physically consistent geometry and appearance.
    (Left) The first row shows a rendered image, normal map, and transparency map. 
    The second row visualizes the radiance contributions of the interface, transmission, and reflection components. (Right) Reconstructed Mesh.
    }
    \label{fig:teaser}
\end{center}

%% file: contents/0_abstract.tex
\begin{abstract}
While 3D Gaussian splatting has emerged as a powerful paradigm, it fundamentally fails to model transparency such as glass panels. The core challenge lies in decoupling the intertwined radiance contributions from transparent interfaces and the transmitted geometry observed through the glass. We present GLINT, a framework that models scene-scale transparency through explicit decomposed Gaussian representation. GLINT reconstructs the primary interface and models reflected and transmitted radiance separately, enabling consistent radiance transport. During optimization, GLINT bootstraps transparency localization from geometry-separation cues induced by the decomposition, together with geometry and material priors from a pre-trained video relighting model. Extensive experiments demonstrate consistent improvements over prior methods for reconstructing complex transparent scenes.
\end{abstract}

%% file: contents/1_intro.tex
\vspace{-16pt}
\section{Introduction}
\label{sec:intro}

Photorealistic 3D reconstruction with accurate geometry is a central goal in vision and graphics. 3D Gaussian Splatting (3DGS)~\cite{kerbl20233d} has recently advanced this goal with real-time rendering and high visual fidelity.
The core challenge in modeling transparent surfaces with 3DGS arises from its monolithic $\alpha$-blending formulation, which inherently conflates geometry and appearance across multiple radiance paths.
Consequently, 3DGS bakes these optical phenomena into entangled Gaussian primitives, achieving visual plausibility while compromising geometric accuracy and interpretability.
The problem is especially pronounced in large-scale scenes containing thin transparent structures, such as architectural glass, display cases, or windows, as in Fig.~\ref{fig:teaser}.
In these scenarios, a single pixel captures a superposition of radiance from reflected and transmitted components originating from distinct physical locations~\cite{wieschollek2018separating}.
To render photorealistic novel view images, the Gaussians associated with glass must have negligibly low opacity or be pruned during optimization to reveal background objects. Conversely, physically grounded glass geometry demands high opacity with sharp boundaries.
This ambiguity forces standard $\alpha$-blending into a compromise, often producing ghost-like reflections or missing transparent geometry.

Existing approaches often rely on object-centric assumptions and segmentation masks~\cite{kim2025transplat, li2025tsgs}, or are restricted to primary first-surface reconstruction, struggling to recover transmissive radiance contributions. Such limitations prevent a physically faithful reconstruction, which is crucial for downstream applications including robotic manipulation, digital-twin construction, and scene understanding.

In this paper, we present \textbf{GLINT}, \textit{a Gaussian Light INverse-rendering framework for scene-scale Transparency reconstruction.}
Specifically, we model a scene with a decomposed Gaussian representation that explicitly partitions primitives into interface, transmission, and reflection components,
enabling a formulation for both transparent and opaque regions. Our hybrid Gaussian radiance transport unifies rasterization and ray tracing to reconstruct multi-path radiance in a physically grounded manner.
In addition, we incorporate geometric and material priors from a pre-trained video diffusion relighting model~\cite{liang2025diffusion}, which complements our decomposition and provides optimization stability.
As illustrated in Fig.~\ref{fig:teaser}, our framework successfully reconstructs complex transparent scenes, producing accurate geometry, reliable transparency maps, and an interpretable decomposition of radiance.
Our key contributions are as follows:
\begin{itemize}
    \item We propose a decomposed Gaussian representation that enables modeling of both transparent and opaque regions through explicit separation of first-visible interface, transmission, and reflection components.
    \item We introduce a hybrid transparency-aware rendering scheme that unifies rasterization and ray tracing under a physically grounded radiance formulation for consistent multi-path radiance transport.
    \item We introduce 3D-FRONT-T, a first synthetic benchmark for scene-scale transparency that enables quantitative evaluation of both appearance and geometry. Using this dataset and the real-world DL3DV-10K~\cite{ling2024dl3dv} dataset, we achieve state-of-the-art reconstruction performance in both photometric and geometric evaluation.
\end{itemize}

%% file: contents/2_related.tex
\section{Related Work}
Our study lies at the intersection of neural scene representations and physically based inverse rendering, which aims to reconstruct accurate geometry and appearance with scene-scale transparency. We review relevant research including: advances in 3DGS, radiance decomposition for non-Lambertian surfaces, and transparent scene reconstruction.

\subsection{Gaussian Splatting}
3DGS~\cite{kerbl20233d} represents scenes using explicit 3D Gaussian primitives, achieving real-time novel view synthesis with high visual fidelity.
Since its introduction, numerous extensions have sought to improve its geometric accuracy and photometric realism.
Several works reduce reliance on strong SfM initialization~\cite{fu2024colmap,kheradmand20243d}, introduce alias-free splatting~\cite{yu2024mip}, or enhance geometric fidelity~\cite{huang20242d,dai2024high,guedon2023sugar,chen2024pgsr}.
For instance, adapting Gaussians into planar-constrained structures~\cite{guedon2023sugar,chen2024pgsr} or surfels~\cite{huang20242d,dai2024high} enables a more faithful representation of fine-grained surface details.

Accurate surface modeling is crucial not only for geometry reconstruction but for capturing secondary lighting effects, such as reflection and refraction, which are sensitive to surface geometry.
However, existing approaches commonly rely on spherical harmonics to approximate view-dependent radiance, which limits their ability to capture complex non-Lambertian effects in an interpretable manner.

\subsection{Radiance Decomposition in Gaussian Splatting}
To move beyond simple view-dependent appearance and model complex light–material interactions, several works decompose radiance into more interpretable components. 

One line of research integrates physically-based material properties directly into the Gaussian primitives. For instance, GaussianShader~\cite{jiang2024gaussianshader} and R3DG~\cite{R3DG2023} incorporate BRDF shading functions for each Gaussian, enabling per-primitive separation into diffuse and specular components.

A different strategy focuses on explicitly modeling the source of reflections rather than just the surface properties. DeferredGS~\cite{wu2024deferredgs} and 3DGS-DR~\cite{ye20243d} employ deferred rendering pipelines, which first buffer geometric properties and then compute view-dependent shading and reflections in a second pass. Taking this concept further, EnvGS~\cite{xie2025envgs} introduces a separate set of environment Gaussians to explicitly model the surrounding scene, then uses a ray tracing pipeline to query this environment representation to render strong reflections.

While these approaches greatly improve the realism and interpretability, their formulations remain restricted to opaque reflected light paths and provide limited capacity for modeling transmitted radiance. Consequently, they struggle with transparent surfaces that exhibit both reflection and transmission, a common case in real-world settings such as glass facades, display cases, and windows.

\subsection{Transparent Scene Reconstruction}
Transparency poses a fundamental challenge for multi-view reconstruction because the observed radiance is a mixture of transmitted background and glass-reflected environment components, violating the single-surface visibility assumption in conventional methods~\cite{wu2025alphasurf}.
For optically thin glass, light transport remains nearly linear with negligible refraction~\cite{gu2007dirty}, yet radiance contributions from multiple depths overlap within each pixel, resulting in perceptually complex observations.
This contrasts with refractive transparency, where pronounced refraction at material interfaces produces explicit optical phenomena that require volumetric modeling of the refractive medium~\cite{li2020through,bemana2022eikonal,sun2024nu}.

Under the Gaussian splatting paradigm, TransparentGS~\cite{huang2025transparentgs} introduces transparent Gaussian primitives and light-field probes to efficiently handle object-centered refraction, while TSGS~\cite{li2025tsgs} targets optically thin transparency using first-surface rasterization for accurate geometry reconstruction. 
Despite these advances, both approaches remain object-centric and require segmentation or assume a single surface, making them difficult to scale to complex multi-depth transparency in real-world scenes.

In this paper, we address these limitations by modeling scene-scale optically thin transparency through a decomposed Gaussian representation and a physically consistent radiance transport formulation that jointly reconstructs geometry and appearance without requiring masks.

%% file: contents/3_method.tex
\input{figures/3_framework}

\section{Method}
\label{sec:method}

\subsection{Preliminaries: 2D Gaussian Rendering}
Our method builds upon Gaussian splatting, specifically adopting 2D Gaussian Splatting (2DGS)~\cite{huang20242d}, which represents scenes using anisotropic 2D Gaussians defined on local tangent planes. This resolves geometric inaccuracies in 3DGS~\cite{kerbl20233d}, which relies on volumetric 3D primitives.

While 3DGS defines primitives with a 3D covariance matrix 
$\boldsymbol{\Sigma} \in \mathbb{R}^{3 \times 3}$, 2DGS constrains each Gaussian’s 
covariance to lie on a local tangent plane. Each 2D Gaussian primitive $G_i$ is 
parameterized by its 3D mean $\boldsymbol{\mu}_i \in \mathbb{R}^3$, a 2D covariance $\boldsymbol{\Sigma}_i^{2D}$ defined on the tangent plane~\cite{huang20242d}, 
Spherical Harmonics (SH) coefficients $\mathbf{c}_i$ for view-dependent color, and an opacity $o_i$.

The scene is rendered using standard alpha-compositing. Unlike 3DGS, which evaluates volumetric density through a projected 2D footprint, 2DGS performs perspective-correct splatting. For each pixel $\mathbf{p}$, the contribution of a primitive is computed by explicitly intersecting the camera ray with the primitive's tangent plane, yielding a local coordinate $\mathbf{u}_i(\mathbf{p})=(u,v)$.
The 2D Gaussian is evaluated as:
\begin{equation}
\label{eq:gauss_eval}
G_i(\mathbf{u}_i) = \exp\!\left[-\tfrac{1}{2}(u^2 + v^2)\right].
\end{equation}
The opacity $\alpha_i(\mathbf{p})$ at which primitive $i$ contributes to pixel $\mathbf{p}$ is given by:
\begin{equation}
\label{eq:2dgs_alpha}
\alpha_i(\mathbf{p}) = o_i \cdot G_i(\mathbf{u}_i(\mathbf{p})).
\end{equation}
With these opacities, the contribution of each primitive along the viewing ray is accumulated using front-to-back compositing. 
We first define the transmittance $T_i$, which measures how much light from primitive $i$ remains unoccluded by all closer primitives $j<i$:
\begin{equation}
\label{eq:transmittance}
T_i = \prod_{j<i} (1-\alpha_j(\mathbf{p})).
\end{equation}
Finally, the pixel radiance is obtained by summing the colors of all primitives intersecting the ray:
\begin{equation}
\label{eq:compositing}
\mathbf{L}(\mathbf{o},\mathbf{d})
=
\sum_{i\in\mathcal{S}(\mathbf{r})}
T_i\,\alpha_i(\mathbf{p})\,\mathbf{c}_i(\mathbf{d}).
\end{equation}
This perspective-correct formulation preserves the correct geometric ordering of splats and enables stable, view-consistent rendering.

\vspace{-3pt}
\paragraph{Rendering techniques.}
In practice, the per-pixel contribution $\alpha_i(\mathbf{p})$ (Eq.~(\ref{eq:2dgs_alpha})) can be computed in two ways.

\noindent
\textit{Rasterization.}
The 2DGS~\cite{huang20242d} framework employs a tile-based rasterizer that identifies affected pixels, computes the perspective-correct ray–splat intersection $\mathbf{u}_i(\mathbf{p})$, evaluates the Gaussian (Eq.~(\ref{eq:gauss_eval})), and composites results via Eq.~(\ref{eq:compositing}). This intersection-based formulation maintains geometric consistency and avoids projection artifacts inherent to the affine projection approximation in~\cite{kerbl20233d}.

\noindent 
\textit{Ray Tracing.}
Alternatively, Gaussian ray tracing~\cite{moenne20243d,xie2025envgs} is employed for secondary effects, as rasterization is impractical for querying unique paths per pixel. Following EnvGS~\cite{xie2025envgs}, we represent each 2D Gaussian as a pair of triangles to enable hardware-accelerated BVH traversal, allowing efficient ray-primitive intersection tests.


\subsection{Decomposed Gaussian Representation}
\label{sec:representation} 

Modeling transparency within standard Gaussian volume rendering is fundamentally limited by its monolithic $\alpha$-blending, which entangles geometry and appearance in a single compositing stream.
This creates an inherent \emph{transparency-depth dilemma}~\cite{li2025tsgs}, where appearance optimization conflicts with geometric accuracy.
When transparent surfaces contribute minimal radiance, optimization drives their opacity toward zero, causing them to vanish under adaptive pruning~\cite{kerbl20233d}.
Conversely, increasing opacity treats them as opaque occluders, blocking transmitted backgrounds and collapsing geometry into a single entangled field.

To resolve this, we introduce a decomposed Gaussian
representation that explicitly partitions the scene's Gaussian 
primitives into three functional sets based on their optical
roles: \textit{Interface} ($\mathcal{G}_{\text{intr}}$), \textit{Transmission} ($\mathcal{G}_{\text{trans}}$), and \textit{Reflection} ($\mathcal{G}_{\text{refl}}$).
The interface component $\mathcal{G}_{\text{intr}}$ captures the primary visible surface encountered by camera rays, encompassing both opaque and transparent material boundaries. 
It serves as the geometric interface, encoding per-point geometry and material attributes that govern how radiance is 
routed through subsequent components. The transmission component $\mathcal{G}_{\text{trans}}$ models background geometry visible through transparent surfaces, capturing transmitted radiance paths.
Finally, the reflection component $\mathcal{G}_{\text{refl}}$ encodes environment radiance reflected at both opaque and transparent interfaces.

\vspace{-15pt}
\paragraph{Hybrid rendering pipeline.}
With this decomposed representation, we employ a hybrid rendering approach that uses rasterization for primary surface visibility and ray tracing for secondary transmission and reflection paths.
The interface component $\mathcal{G}_{\text{intr}}$ is rasterized to produce a G-buffer 
$\mathcal{B} = \{z, \mathbf{n}, t, s\}$ encoding depth, normal, transparency, and specularity, 
where each entry is obtained by applying the compositing operator 
$\mathcal{A}(\cdot)$ with weights $\mathcal{A}_i = T_i \alpha_i$ defined in Eq.~\ref{eq:compositing}.
This G-buffer then guides ray-traced queries into $\mathcal{G}_{\text{trans}}$ and $\mathcal{G}_{\text{refl}}$ for secondary transmission and reflection, where the transparency $t \in [0,1]$ gates the opaque–transparent split and the specularity $s \in [0,1]$ controls the diffuse–specular balance.
The overall rendering pipeline is illustrated in Fig.~\ref{fig:framework}.

\subsection{Transparency-Aware Radiance Transport}
\label{sec:radiance_transport}
In this section, we formulate radiance transport through our decomposed Gaussian representation, where outgoing radiance is computed by querying and combining the radiance from interface, transmission, and reflection components based on local surface properties.
Unlike prior radiance decomposition approaches~\cite{wu2024deferredgs,ye20243d,
zhang2025ref,xie2025envgs} which primarily address opaque reflection, our 
formulation adopts a BSDF-inspired decomposition~\cite{bartell1981theory} that splits outgoing 
radiance into reflection and transmission paths based on the surface 
properties encoded in the G-buffer.

The outgoing radiance, which we denote as $L_o$, is expressed as a transparency-gated interpolation between two transport branches:
\begin{equation}
\label{eq:main_interp}
L_o = (1-t)\,L_{\text{opaque}} + t\,L_{\text{transparent}},
\end{equation}
where transparency $t$ obtained from the G-buffer $\mathcal{B}$ determines whether radiance transport follows opaque or transparent paths.

\vspace{-5pt}
\paragraph{Opaque branch ($L_{\text{opaque}}$).}
For opaque surfaces, the outgoing radiance consists of the interface base color combined with reflected environment radiance, weighted by surface specularity.
The transport is modeled as a physically-inspired blend between diffuse and specular components.
The Fresnel reflectance is approximated using the Schlick formulation~\cite{schlick1994inexpensive} with the outgoing direction $\omega_o = -\mathbf{d}$:
\begin{equation}
\label{eq:schlick_noeta}
F(\omega_o) = F_0 + (1 - F_0)(1 - \max(0,\,\omega_o \!\cdot\! \mathbf{n}))^5,
\end{equation}
where $F_0$ denotes the normal-incidence reflectance.
A learnable per-pixel specularity $s$ modulates the overall specular weight, yielding a Fresnel-weighted blending factor:
\begin{equation}
\label{eq:spec_ks}
k_s = s + (1 - s)F(\omega_o),
\end{equation}
and the outgoing opaque radiance is given by:
\begin{equation}
\label{eq:opaque_rad}
L_{\text{opaque}} = (1 - k_s)\,L_{\text{intr}} + k_s\,L_{\text{refl}},
\end{equation}
where $L_{\text{intr}}$ is the base color rasterized from the interface set $\mathcal{G}_{\text{intr}}$, and
$L_{\text{refl}}(\mathbf{x}, \omega_r) = \operatorname{Trace}(\mathcal{G}_{\text{refl}}, \mathbf{x}, \omega_r)$
denotes radiance traced from the reflection component $\mathcal{G}_{\text{refl}}$ along the analytic reflection direction:
\begin{equation}
\label{eq:reflection_dir}
\omega_r = 2(\mathbf{n} \!\cdot\! \omega_o)\,\mathbf{n} - \omega_o.
\end{equation}
This approximates the behavior of a diffuse–specular BRDF while omitting roughness and masking–shadowing terms for simplicity.

\vspace{-6pt}
\paragraph{Transparent branch ($L_{\text{transparent}}$).}
For transparent surfaces, the outgoing radiance consists of transmitted radiance from the background and reflected radiance from the environment.
We adopt a Fresnel-based decomposition inspired by the dielectric BSDF to balance these two contributions.
The outgoing transparent radiance is formulated as:
\begin{equation}
\label{eq:transparent_rad}
L_{\text{transparent}} = (1 - k_s)\,L_{\text{trans}} + k_s\,L_{\text{refl}},
\end{equation}
where $k_s$ is derived from Eq.~\ref{eq:spec_ks}, modulating the transmitted and reflected contributions.
Under the optically thin assumption ($\omega_t \approx \omega_o$), refraction-induced 
bending is negligible, allowing the transmitted radiance to be approximated by 
$L_{\text{trans}} = \operatorname{Trace}(\mathcal{G}_{\text{trans}}, \mathbf{x}, \omega_t)$.

By separating transmission and reflection into independently optimizable Gaussian components, this formulation mitigates the transparency–depth trade-off inherent in monolithic $\alpha$-blending, which often leads to degraded visual fidelity or geometric accuracy.

\input{figures/layered_depths}

\subsection{Optimization}
\label{sec:training}
Our primary training objective minimizes the photometric error between the rendered outgoing radiance $L_o$ (Eq.~\ref{eq:main_interp}) and the ground-truth image $L_{\text{GT}}$:
\begin{equation}
\label{eq:photo_loss}
\mathcal{L}_{\text{photo}} = \lambda_1 \mathcal{L}_1 + \lambda_{\text{ssim}} \mathcal{L}_{\text{SSIM}} + \lambda_{\text{lpips}} \mathcal{L}_{\text{LPIPS}},
\end{equation}
where $\mathcal{L}_1$ measures pixel-wise reconstruction error, $\mathcal{L}_{\text{SSIM}}$~\cite{wang2004image} enforces structural similarity, and $\mathcal{L}_{\text{LPIPS}}$~\cite{zhang2018unreasonable} captures perceptual fidelity.

While $\mathcal{L}_{\text{photo}}$ provides the main supervision signal, transparent scenes remain ill-posed because each pixel mixes reflected and transmitted radiance from different depths, making photometric cues alone insufficient. Prior works use dense monocular predictors~\cite{depth_anything_v2,ye2024stablenormal,ren2024grounded} as auxiliary geometric regularizers~\cite{li2024dngaussian,xie2025envgs,li2025tsgs}.
Instead, we utilize the encoder of a pre-trained video relighting model~\cite{liang2025diffusion} to obtain frame-consistent geometric and material priors that regularize the interface components.

\input{figures/trans_map}

\vspace{-6pt}
\paragraph{Geometric regularization.}
Specifically, we use the predicted depth $\hat{z}$ and normal $\hat{\mathbf{n}}$ from the encoder~\cite{liang2025diffusion} to regularize the interface component's geometric attributes in G-buffer:
\begin{equation}
\mathcal{L}_{\text{geo}} =
\lambda_d \mathcal{L}_{\text{depth}}(z, \hat{z})
+ \lambda_n \mathcal{L}_{\text{normal}}(\textbf{n}, \hat{\textbf{n}}),
\end{equation}
where $\mathcal{L}_{\text{depth}}$ follows a scale-invariant formulation~\cite{yu2022monosdf}, and $\mathcal{L}_{\text{normal}} = 1 - \cos(\textbf{n}, \hat{\textbf{n}})$ penalizes angular deviation~\cite{li2025tsgs}.
These priors stabilize the interface geometry, providing a robust scaffold for the secondary transmission and reflection components.

\input{tables/Table_2_geometry}
\input{figures/qual_3d_front_t}
\input{figures/qual_mesh}

\input{figures/qual_dl3dv}

\vspace{-6pt}
\paragraph{Bootstrapping transparency.}
A key advantage of our decomposed representation is its ability to bootstrap transparency localization without manual segmentation masks.
While segmentation modules~\cite{ren2024grounded} can identify 
transparent objects (\ie, glass, bottle) in isolation, they often 
fail or return noisy results for scene-scale transparency due to 
ambiguous boundaries and overlapping transmitted radiance.

Instead, our decomposed representation bootstraps transparency localization, leveraging signals that emerge during optimization. 
The principal cue is the inter-component depth difference $\Delta z = |z_{\text{intr}} - z_{\text{trans}}|$, where $z_{\text{intr}}$ and $z_{\text{trans}}$ denote the interface and transmitted depths respectively. 
This signal, arising where multiple depth layers coexist, reveals the spatial separation between the interface and transmitted geometry as in Fig.~\ref{fig:layered_depth}. 
As a complementary signal, we use the diffuse-albedo map $\hat{a}$ obtained from~\cite{liang2025diffusion}, where lower values help identify specular-dominant transport rather than diffuse transport.
We construct a binary transparency mask by thresholding these bootstrapped cues:
\begin{equation}
M_{\text{trans}} =
\mathbf{1}\bigl((\Delta z > \tau_d) \land (\hat{a} < \gamma_a)\bigr),
\end{equation}
which supervises the predicted transparency buffer $t$ via:
\begin{equation}
\mathcal{L}_{\text{trans}} = \lambda_t \| M_{\text{trans}} - t \|_1.
\end{equation}
As shown in Fig.~\ref{fig:transmap}, this bootstrapped transparency 
localization effectively identifies glass regions across diverse scenes, exhibiting inherent transparency disentanglement due to our explicit decomposed representation.

%% file: figures/3_framework.tex
\begin{figure*}[ht!]
    \centering
    \includegraphics[width=1.0\linewidth,trim=0 6.6cm 0.38cm 0,clip]{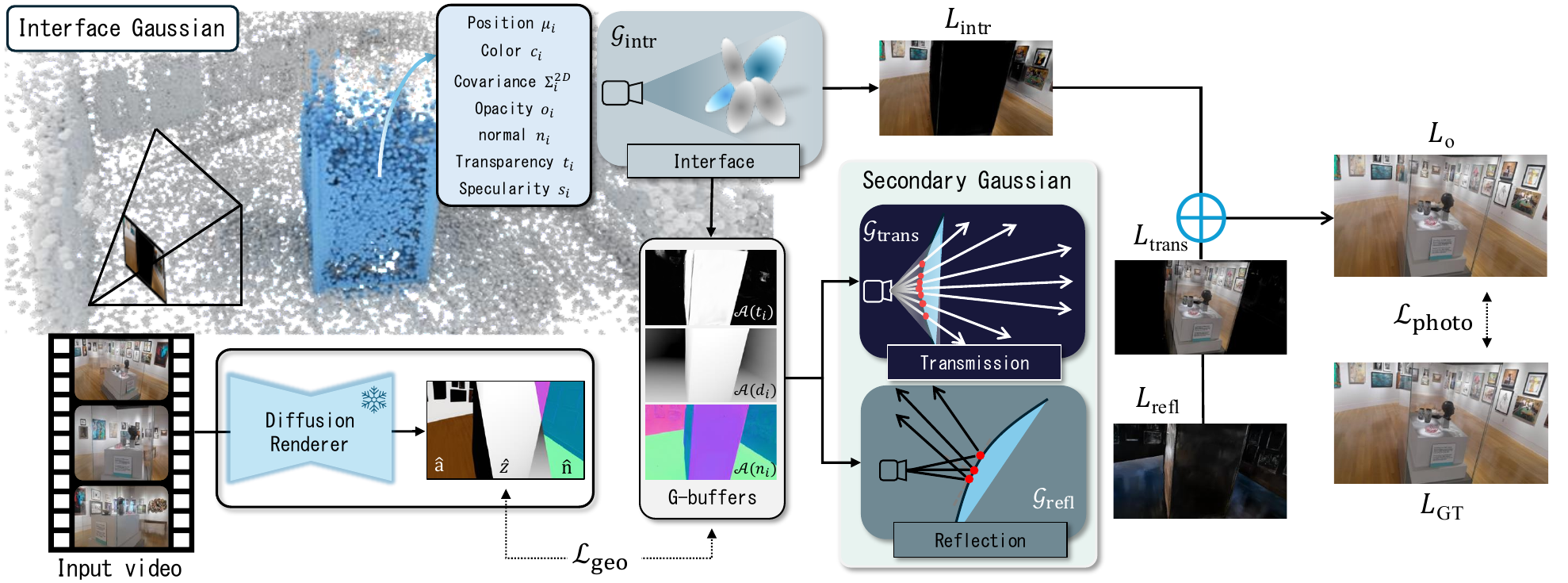}
    \caption{
        \textbf{Pipeline Overview.} 
        The interface component $\mathcal{G}_{\text{intr}}$ captures primary first-surface, while secondary components $\mathcal{G}_{\text{trans}}$ and $\mathcal{G}_{\text{refl}}$ separately model transmission and reflection.
        The output color $L_o$ is obtained through hybrid rendering under transparency-aware radiance transport and supervised by photometric loss $\mathcal{L}_{\text{photo}}$.
        DiffusionRenderer~\cite{liang2025diffusion} provides priors that regularize the G-buffers via $\mathcal{L}_{\text{geo}}$.
        }
        \label{fig:framework}
    \label{fig:framework}
    \vspace{-10pt}
\end{figure*}

%% file: figures/layered_depths.tex
\begin{figure}[t!]
    \centering

    \begin{subfigure}[b]{0.32\linewidth}
        \centering
        \includegraphics[width=\linewidth]{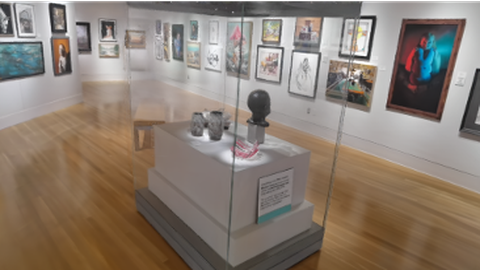}
    \end{subfigure}
    \begin{subfigure}[b]{0.32\linewidth}
        \centering
        \includegraphics[width=\linewidth]{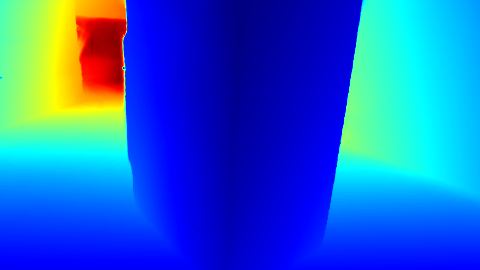}
    \end{subfigure}
    \begin{subfigure}[b]{0.32\linewidth}
        \centering
        \includegraphics[width=\linewidth]{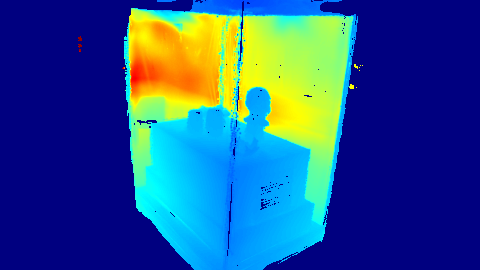}
    \end{subfigure}

    \vspace{2pt}

    \begin{subfigure}[b]{0.32\linewidth}
        \centering
        \includegraphics[width=\linewidth]{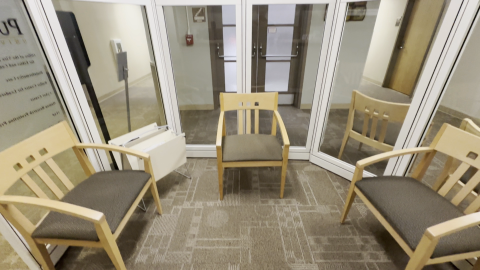}
    \end{subfigure}
    \begin{subfigure}[b]{0.32\linewidth}
        \centering
        \includegraphics[width=\linewidth]{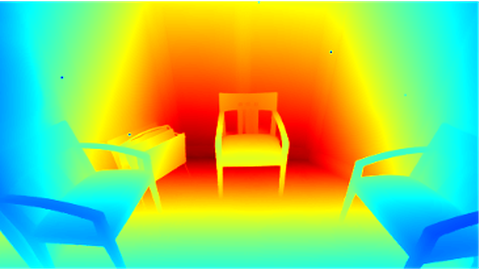}
    \end{subfigure}
    \begin{subfigure}[b]{0.32\linewidth}
        \centering
        \includegraphics[width=\linewidth]{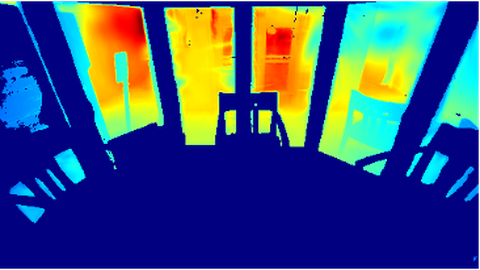}
    \end{subfigure}

    \vspace{-8pt}
    \caption{
        \textbf{Depth decomposition.} (Left) Rendered image, (Middle) interface depth, and (Right) transmission depth.
    }
    \vspace{-5pt}
    \label{fig:layered_depth}
\end{figure}

%% file: figures/trans_map.tex
\begin{figure}[t!]
    \centering
    \includegraphics[width=0.246\linewidth]{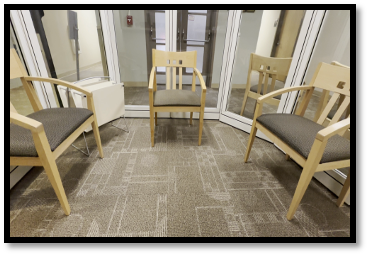}\hspace{-1.5pt}%
    \includegraphics[width=0.246\linewidth]{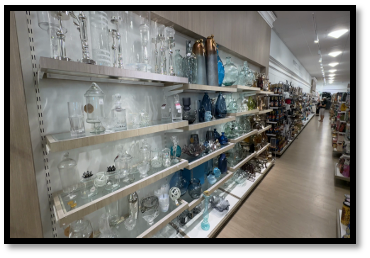}\hspace{-1.5pt}%
    \includegraphics[width=0.246\linewidth]{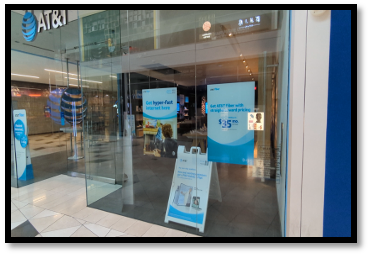}\hspace{-1.5pt}%
    \includegraphics[width=0.246\linewidth]{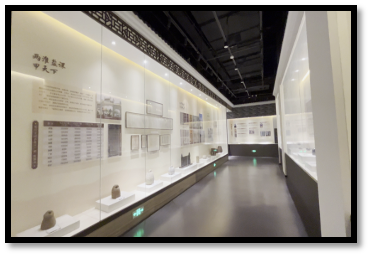}
    \\[1pt]
    \includegraphics[width=0.246\linewidth]{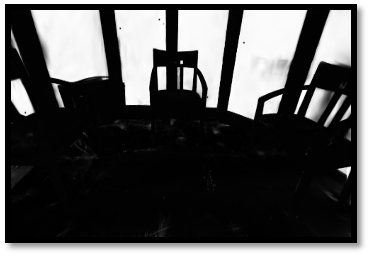}\hspace{-1.5pt}%
    \includegraphics[width=0.246\linewidth]{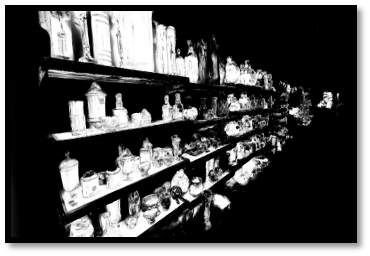}\hspace{-1.5pt}%
    \includegraphics[width=0.246\linewidth]{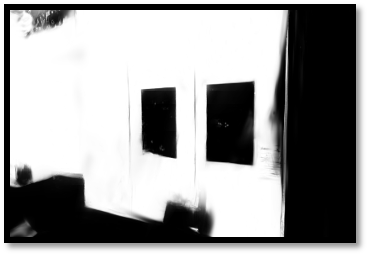}\hspace{-1.5pt}%
    \includegraphics[width=0.246\linewidth]{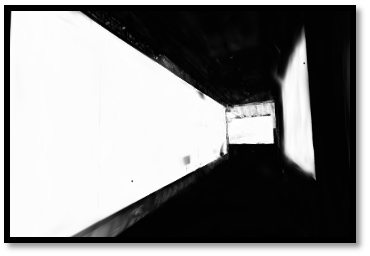}

    \vspace{-5pt}
    \caption{
        \textbf{Obtained transparency maps.}
        GT images (first row) and learned transparency maps (second row).
    }
    \label{fig:transmap}
    \vspace{-8pt}
\end{figure}

%% file: tables/Table_2_geometry.tex
\begin{table*}[t!]
\centering
\scriptsize
\caption{
\textbf{Quantitative evaluation of geometry on the synthetic 3D-FRONT-T dataset.}
We report normal metrics (MAE, and accuracy thresholds of $11.25^\circ$, $22.5^\circ$) 
and depth metrics (AbsRel, RMSE, $\delta<1.25$), along with mesh metrics (CD, F1-score). 
}
\resizebox{0.95\textwidth}{!}{
\begin{tabular}{lcccccccc}
\toprule
\multirow{2}{*}{Method} &
\multicolumn{3}{c}{Normal} &
\multicolumn{3}{c}{Depth} &
\multicolumn{2}{c}{Mesh} \\
\cmidrule(lr){2-4} \cmidrule(lr){5-7} \cmidrule(lr){8-9}
 & MAE$\downarrow$ & $11.25^\circ\uparrow$ & $22.5^\circ\uparrow$ &
AbsRel$\downarrow$ & RMSE$\downarrow$ & $\delta<1.25\uparrow$ &
CD$\downarrow$ & F1$\uparrow$ \\
\midrule
2DGS~\citep{huang20242d}     
& 25.97 & 52.19 & 64.55 
& 0.20 & 0.24 & 76.71 
& \cellthird0.85 & 0.688 \\

PGSR~\citep{chen2024pgsr}    
& 25.39 & 56.83 & 65.58 
& 0.22 & 0.25 & 76.68 
& \cellsecond0.52 & \cellsecond0.807 \\

Ref-GS~\citep{zhang2025ref}  
& 41.55 & 18.61 & 36.79 
& 0.28 & 0.36 & 57.52 
& 1.29 & 0.408 \\

EnvGS~\citep{xie2025envgs}   
& \cellthird14.37 & \cellthird68.22 & \cellthird80.23
& \cellthird0.13 & \cellthird0.16 & \cellthird86.10
& 0.87 & 0.640 \\

TSGS~\citep{li2025tsgs}      
& \cellsecond9.89 & \cellsecond86.29 & \cellsecond92.24
& \cellsecond0.08 & \cellsecond0.12 & \cellsecond95.56
& \cellsecond0.52 & \cellthird0.798 \\

\textbf{GLINT (Ours)}        
& \cellfirst7.96 & \cellfirst86.37 & \cellfirst92.28
& \cellfirst0.04 & \cellfirst0.07 & \cellfirst98.32
& \cellfirst0.34 & \cellfirst0.836 \\

\bottomrule
\end{tabular}
}
\label{tab:geo}
\end{table*}

%% file: figures/qual_3d_front_t.tex
\begin{figure*}[t!]
\centering
\setlength{\tabcolsep}{1pt}

\begin{center}
\renewcommand{\arraystretch}{0.9}
\begin{tabular}{ccccc}
\makebox[0.19\linewidth][c]{{\textrm{GT}}} &
\makebox[0.19\linewidth][c]{{\textrm{PGSR}}~\cite{chen2024pgsr}} &
\makebox[0.19\linewidth][c]{{\textrm{EnvGS}}~\cite{xie2025envgs}} &
\makebox[0.19\linewidth][c]{{\textrm{TSGS}}~\cite{li2025tsgs}} &
\makebox[0.19\linewidth][c]{{\textrm{Ours}}} \\
\end{tabular}
\end{center}

\vspace{-7pt}

\begin{subfigure}[t]{0.95\linewidth}
\centering
\includegraphics[width=0.195\linewidth]{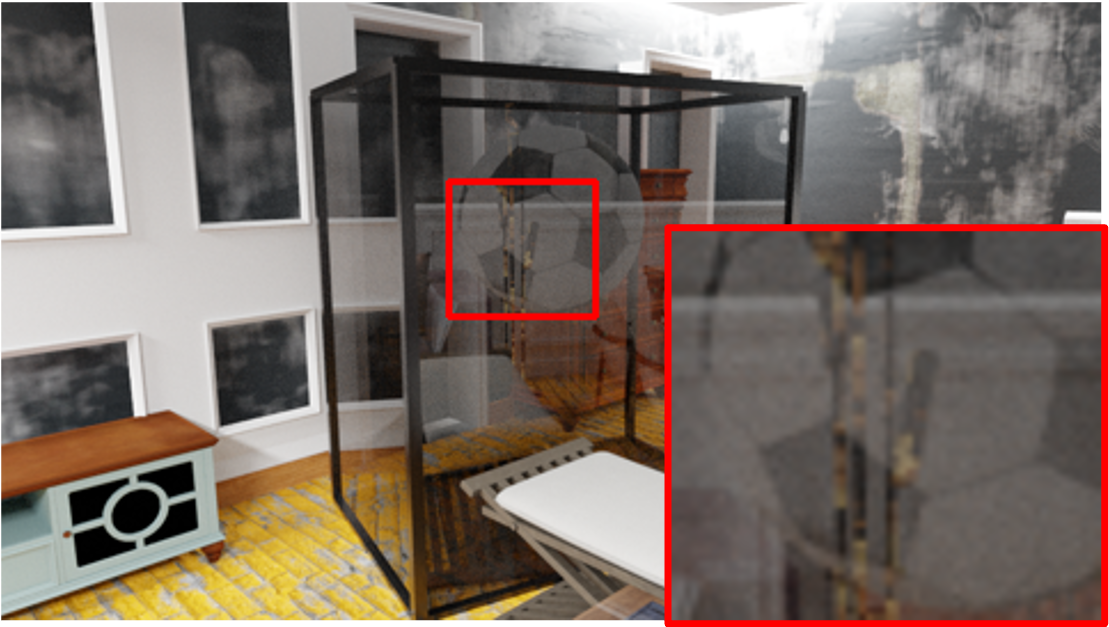}\hspace{1pt}%
\includegraphics[width=0.195\linewidth]{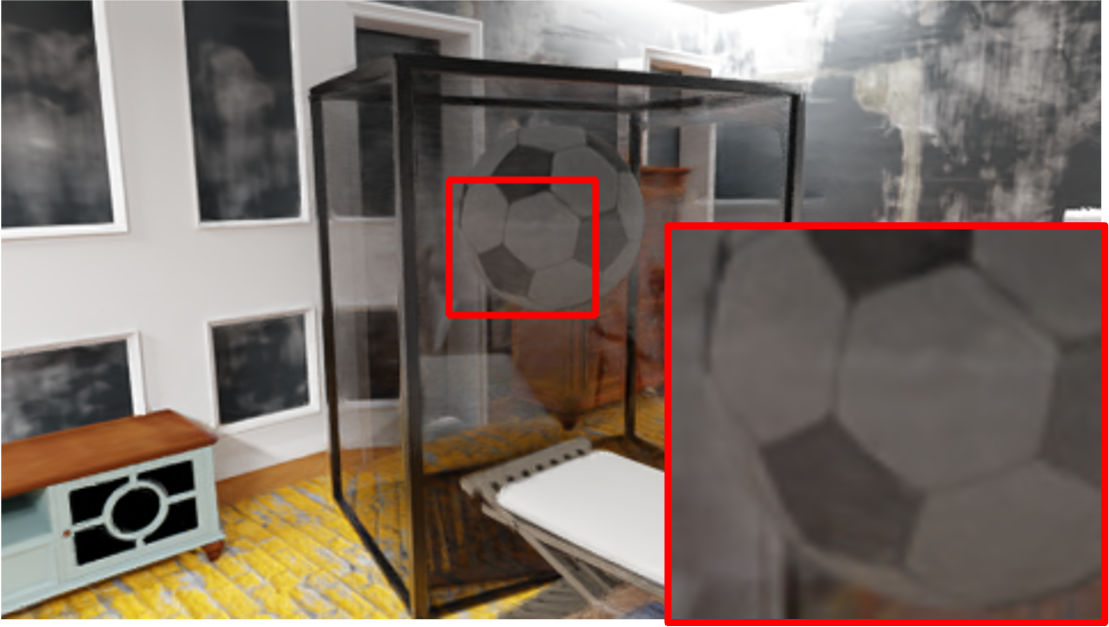}\hspace{1pt}%
\includegraphics[width=0.195\linewidth]{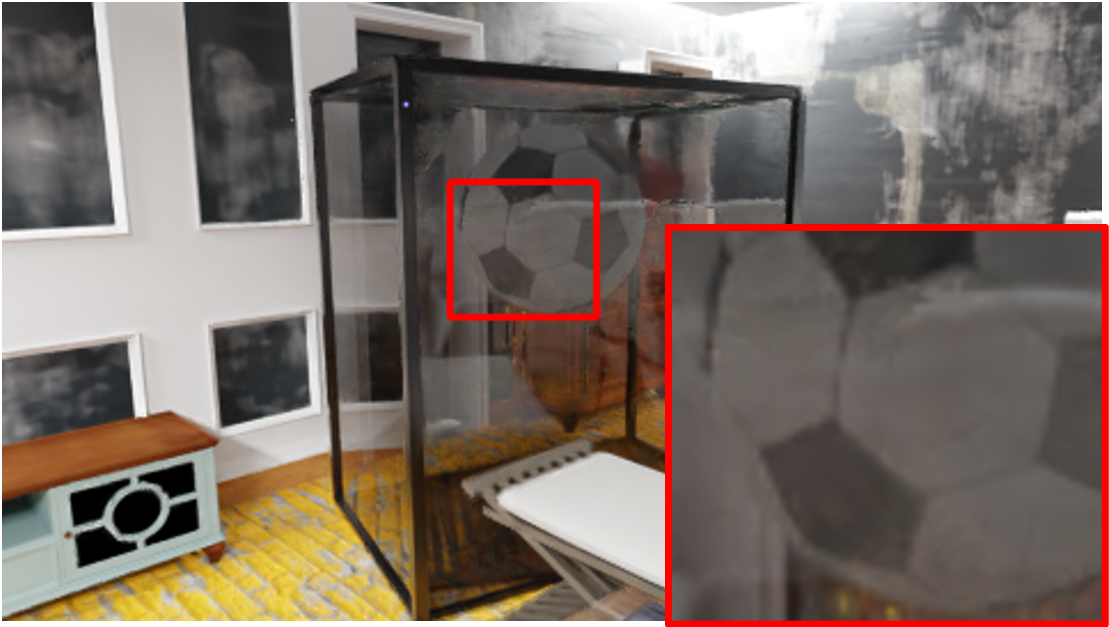}\hspace{1pt}%
\includegraphics[width=0.195\linewidth]{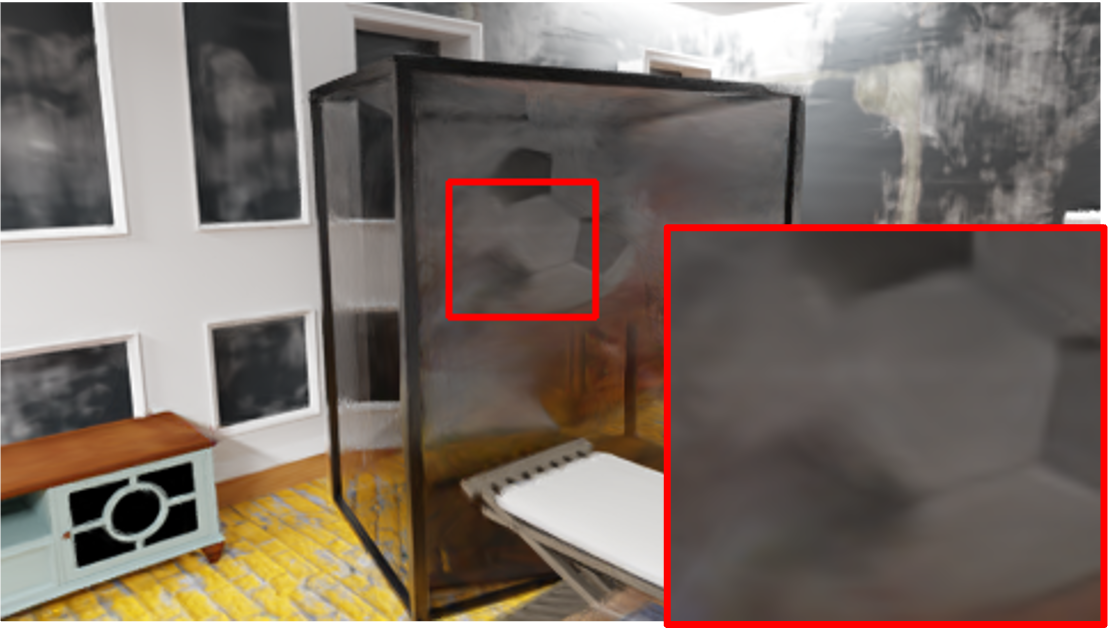}\hspace{1pt}%
\includegraphics[width=0.195\linewidth]{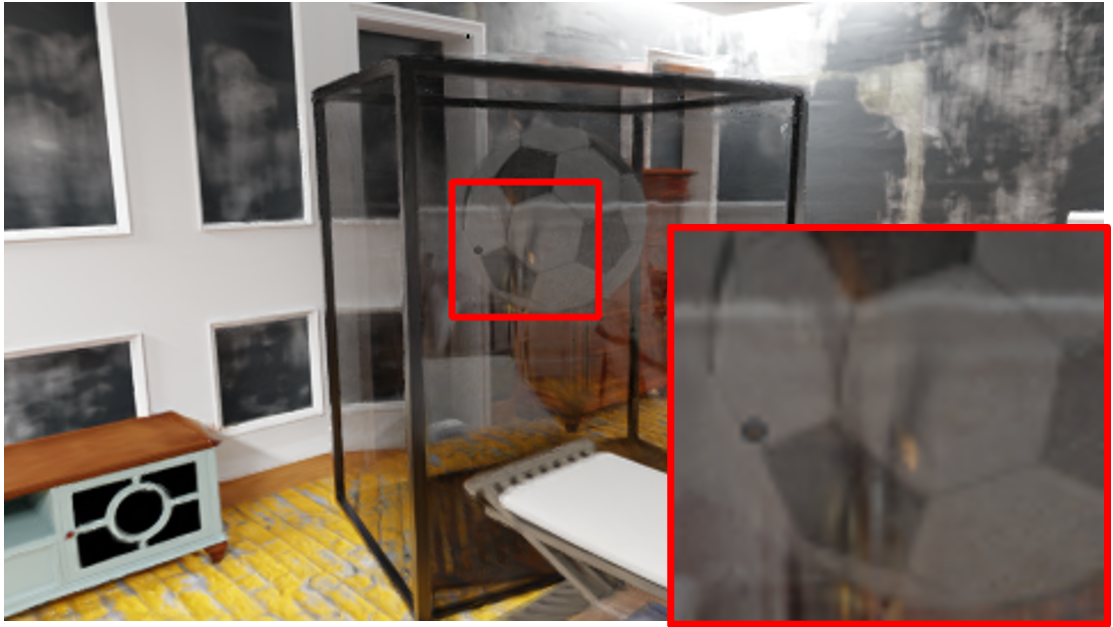}\\[1pt]

\includegraphics[width=0.195\linewidth]{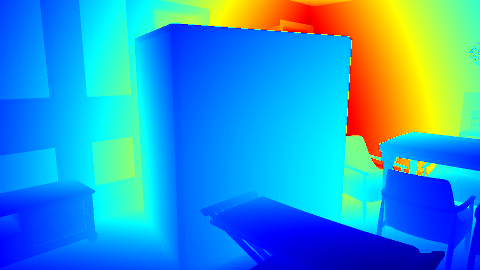}\hspace{1pt}%
\includegraphics[width=0.195\linewidth]{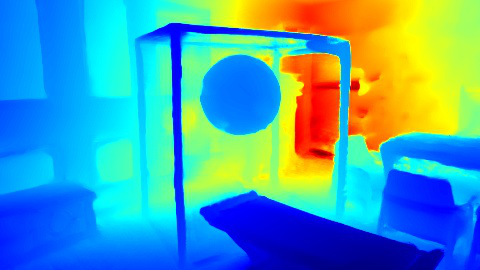}\hspace{1pt}%
\includegraphics[width=0.195\linewidth]{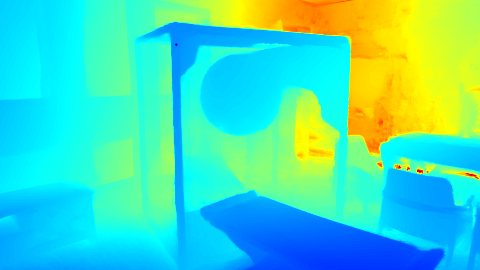}\hspace{1pt}%
\includegraphics[width=0.195\linewidth]{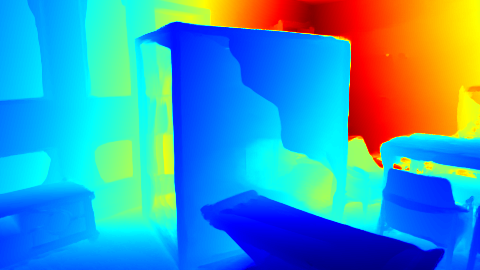}\hspace{1pt}%
\includegraphics[width=0.195\linewidth]{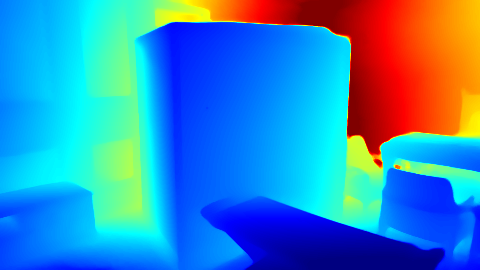}\\[1pt]


\end{subfigure}

\vspace{2pt}\hrule\vspace{2pt}

\begin{subfigure}[t]{0.95\linewidth}
\centering
\includegraphics[width=0.195\linewidth]{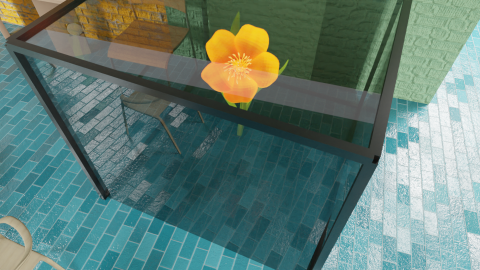}\hspace{1pt}%
\includegraphics[width=0.195\linewidth]{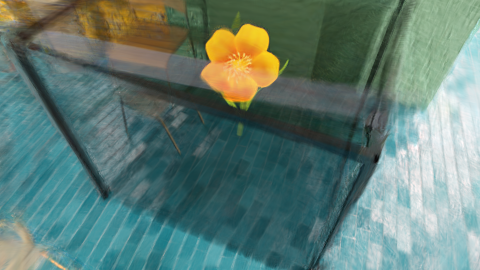}\hspace{1pt}%
\includegraphics[width=0.195\linewidth]{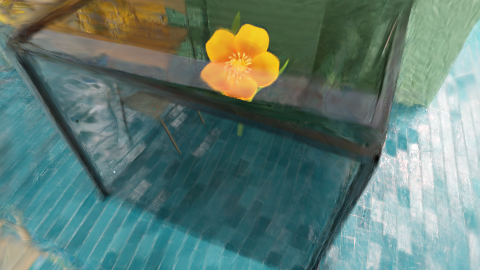}\hspace{1pt}%
\includegraphics[width=0.195\linewidth]{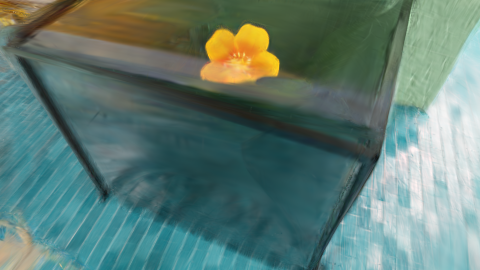}\hspace{1pt}%
\includegraphics[width=0.195\linewidth]{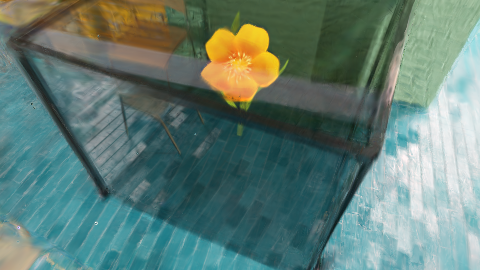}\\[1pt]

\includegraphics[width=0.195\linewidth]{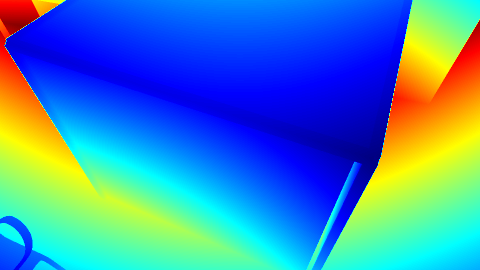}\hspace{1pt}%
\includegraphics[width=0.195\linewidth]{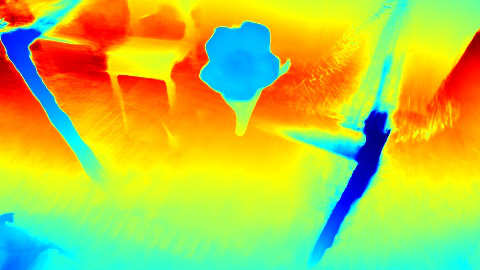}\hspace{1pt}%
\includegraphics[width=0.195\linewidth]{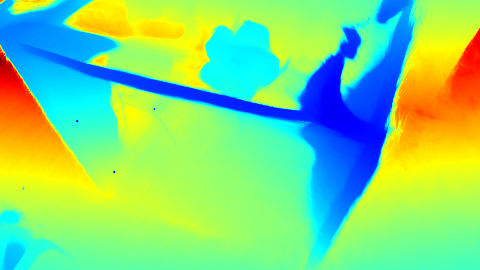}\hspace{1pt}%
\includegraphics[width=0.195\linewidth]{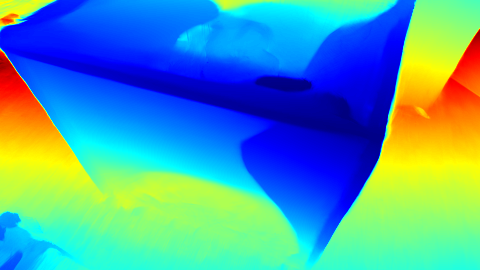}\hspace{1pt}%
\includegraphics[width=0.195\linewidth]{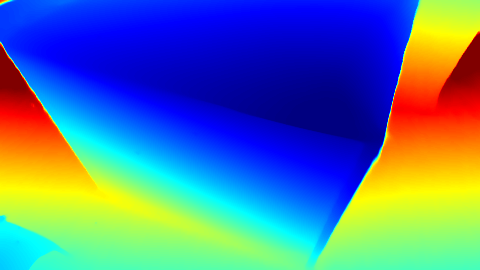}\\[1pt]


\end{subfigure}

\vspace{-3pt}
\caption{
\textbf{Qualitative comparison on synthetic scenes.}
Each column shows results from GT, PGSR~\cite{chen2024pgsr}, EnvGS~\cite{xie2025envgs}, TSGS~\cite{li2025tsgs}, and Ours.
For each scene, rows correspond to RGB (top) and depth maps (bottom).
}
\vspace{-9pt}
\label{fig:qual_synthetic}
\end{figure*}

%% file: figures/qual_mesh.tex
\begin{figure}[t!]
    \centering

    \begin{subfigure}[b]{0.32\linewidth}
        \centering
        \includegraphics[width=\linewidth]{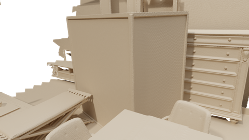}
        \\ GT mesh
    \end{subfigure}
    \hfill
    \begin{subfigure}[b]{0.32\linewidth}
        \centering
        \includegraphics[width=\linewidth]{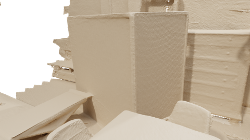}
        \\ Ours
    \end{subfigure}
    \hfill
    \begin{subfigure}[b]{0.32\linewidth}
        \centering
        \includegraphics[width=\linewidth]{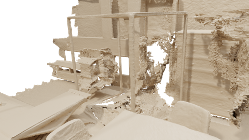}
        \\ EnvGS~\cite{xie2025envgs}
    \end{subfigure}

    \vspace{1pt} 

    \begin{subfigure}[b]{0.32\linewidth}
        \centering
        \includegraphics[width=\linewidth]{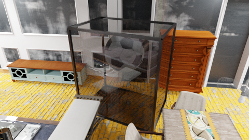}
        \\ RGB
    \end{subfigure}
    \hfill
    \begin{subfigure}[b]{0.32\linewidth}
        \centering
        \includegraphics[width=\linewidth]{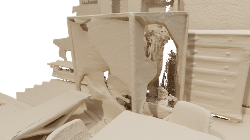}
        \\ TSGS~\cite{li2025tsgs}
    \end{subfigure}
    \hfill
    \begin{subfigure}[b]{0.32\linewidth}
        \centering
        \includegraphics[width=\linewidth]{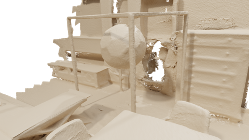}
        \\PGSR~\cite{chen2024pgsr}
    \end{subfigure}

    \vspace{-5pt} 
    \caption{
        \textbf{Mesh visualization comparison.} The meshes are obtained from TSDF fusion following baselines.
    }
    \vspace{-15pt} 
    \label{fig:mesh_visualization}
\end{figure}

%% file: figures/qual_dl3dv.tex
\begin{figure*}[t!]
    \centering
    \setlength{\tabcolsep}{1pt}
    
    \begin{center}
    \renewcommand{\arraystretch}{0.88}
    \begin{tabular}{ccccc}
        \makebox[0.19\linewidth][c]{{\textrm{GT}}} & 
        \makebox[0.19\linewidth][c]{{\textrm{PGSR}}~\cite{chen2024pgsr}} & 
        \makebox[0.19\linewidth][c]{{\textrm{EnvGS}}~\cite{xie2025envgs}} & 
        \makebox[0.19\linewidth][c]{{\textrm{TSGS}}~\cite{li2025tsgs}} & 
        \makebox[0.19\linewidth][c]{{\textrm{Ours}}}
    \end{tabular}
    \end{center}
    \vspace{-7pt}
    
    \begin{subfigure}[t]{\linewidth}
        \centering
        \includegraphics[width=0.195\linewidth]{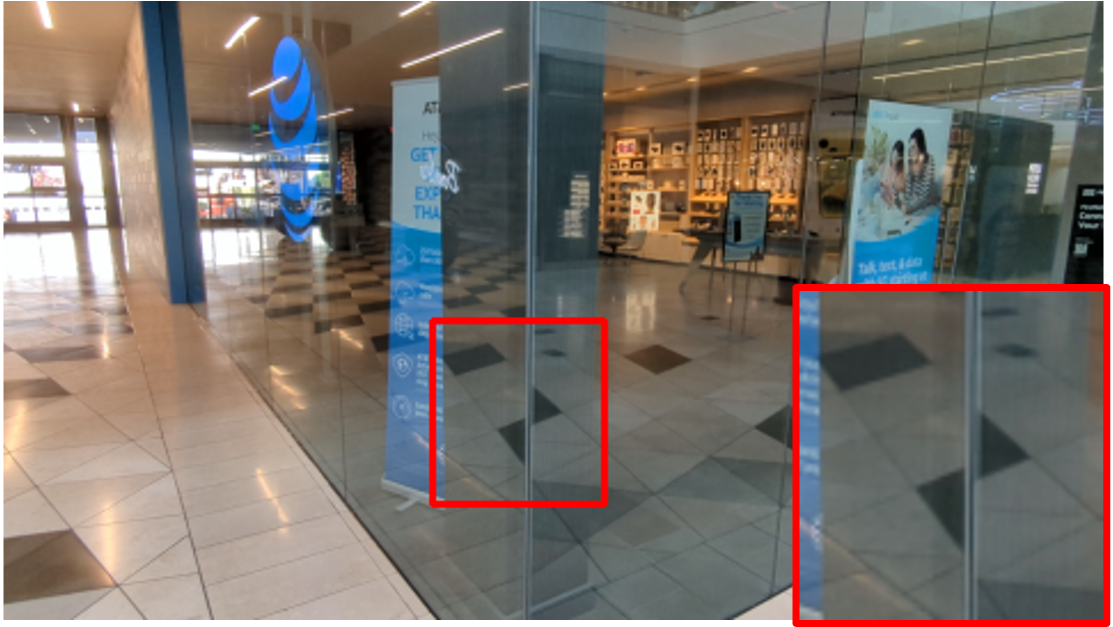}\hspace{1pt}%
        \includegraphics[width=0.195\linewidth]{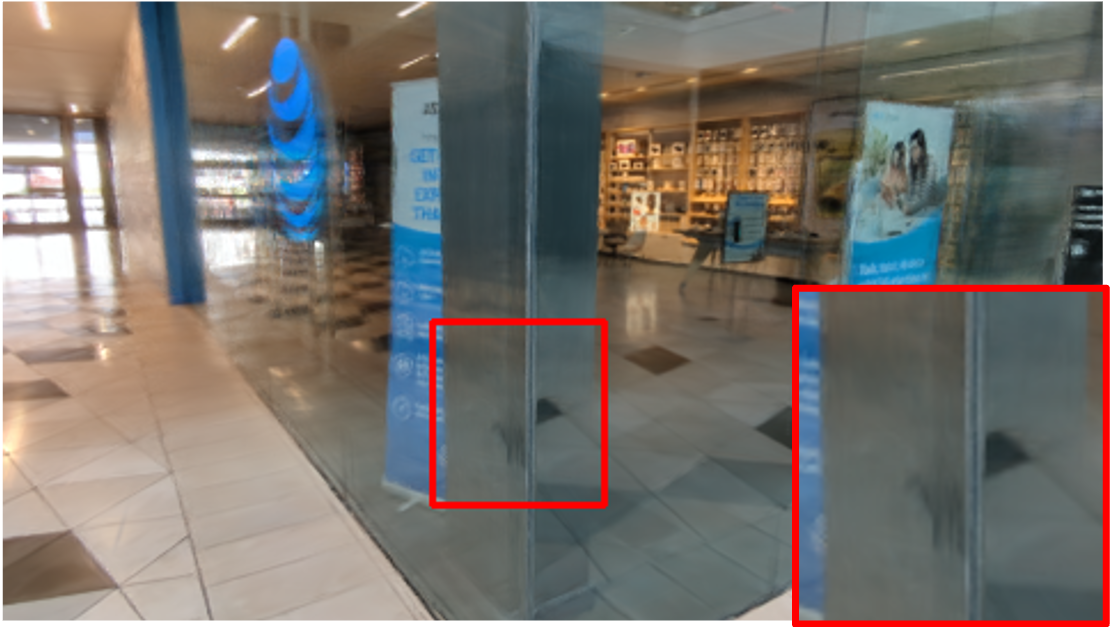}\hspace{1pt}%
        \includegraphics[width=0.195\linewidth]{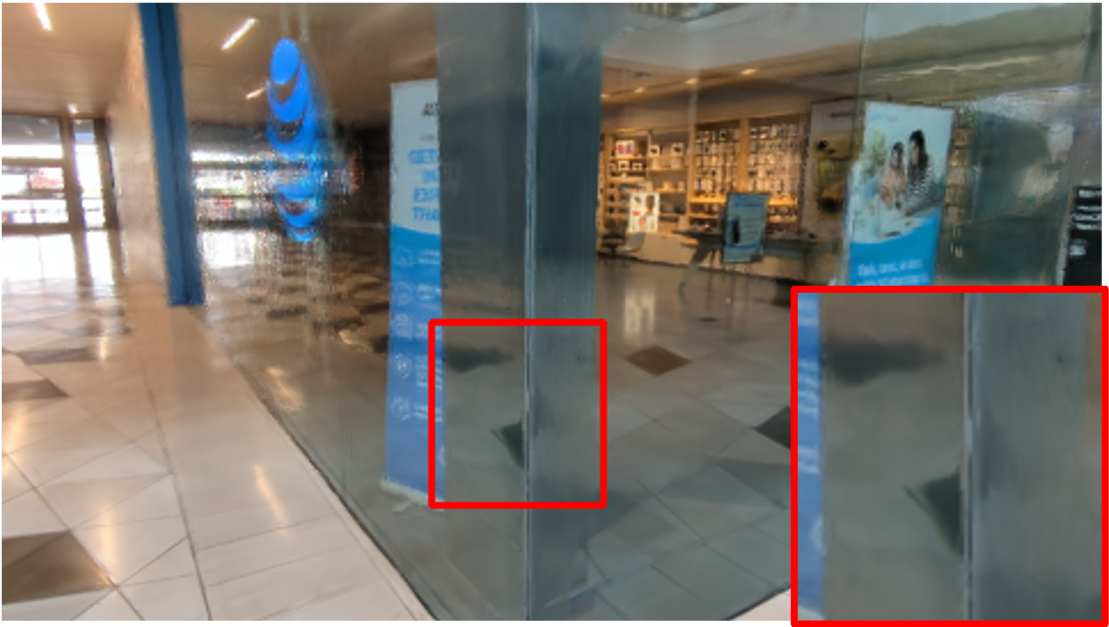}\hspace{1pt}%
        \includegraphics[width=0.195\linewidth]{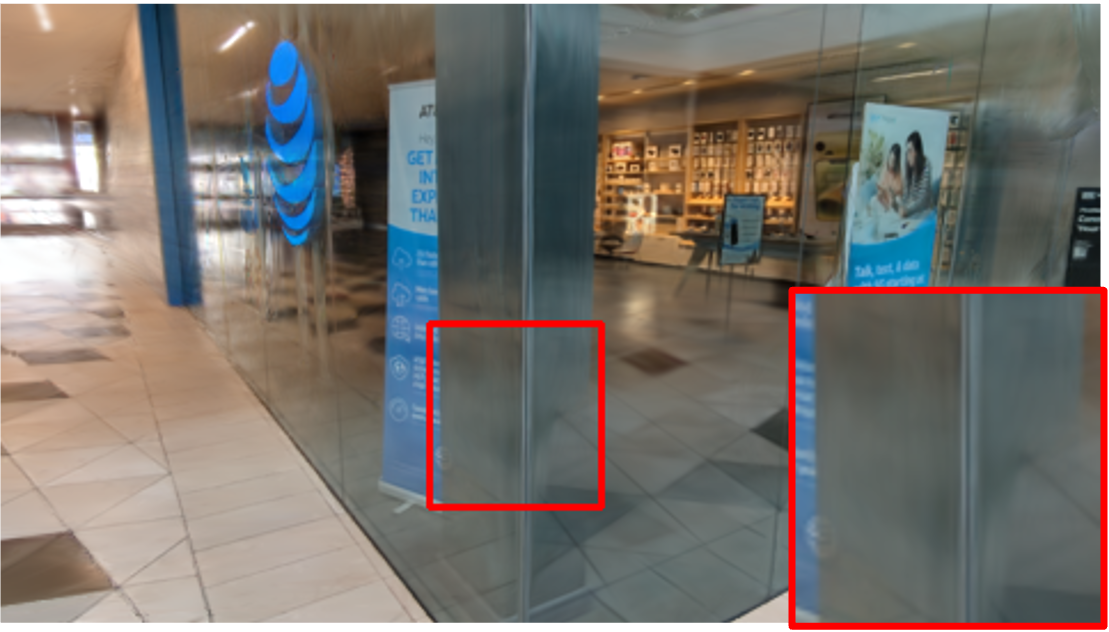}\hspace{1pt}%
        \includegraphics[width=0.195\linewidth]{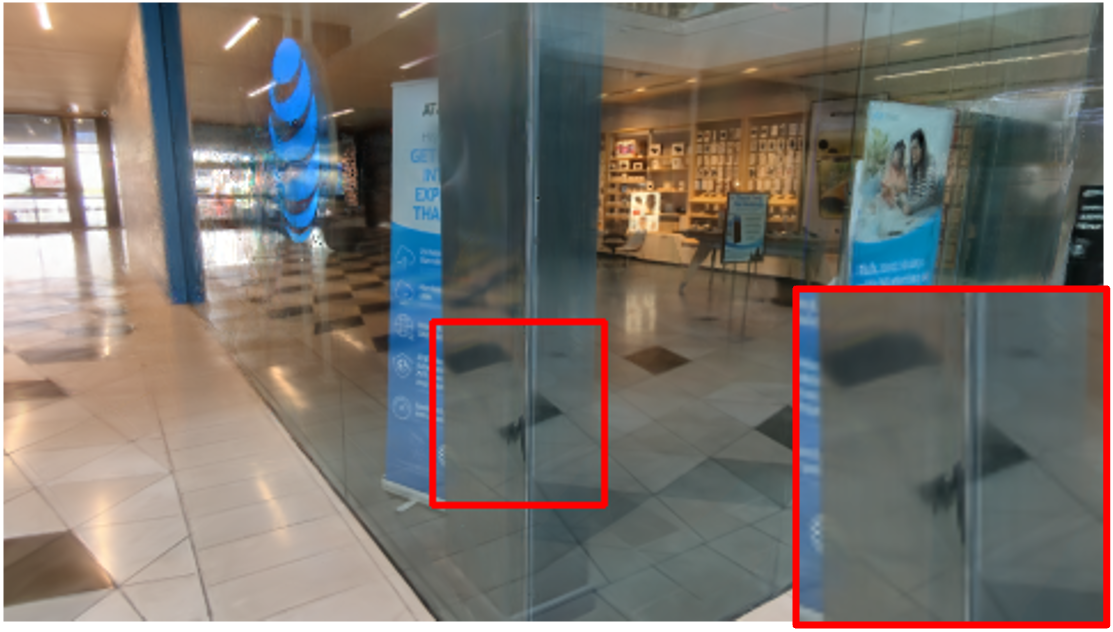}\\[1pt]
        \hspace{0.195\linewidth}\hspace{1pt}%
        \includegraphics[width=0.195\linewidth]{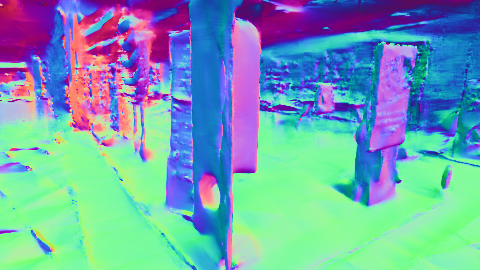}\hspace{1pt}%
        \includegraphics[width=0.195\linewidth]{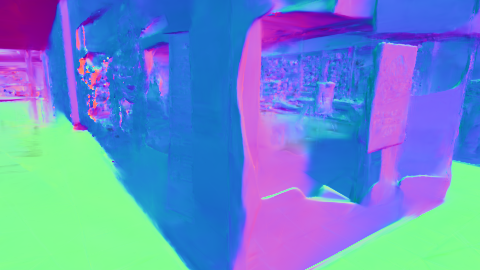}\hspace{1pt}%
        \includegraphics[width=0.195\linewidth]{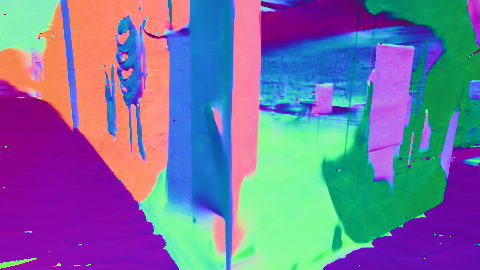}\hspace{1pt}%
        \includegraphics[width=0.195\linewidth]{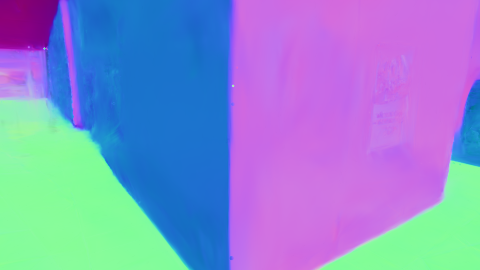}
    \end{subfigure}
    
    \vspace{3pt}
    
    \begin{subfigure}[t]{\linewidth}
        \centering
        \includegraphics[width=0.195\linewidth]{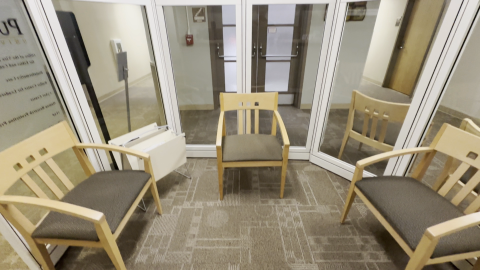}\hspace{1pt}%
        \includegraphics[width=0.195\linewidth]{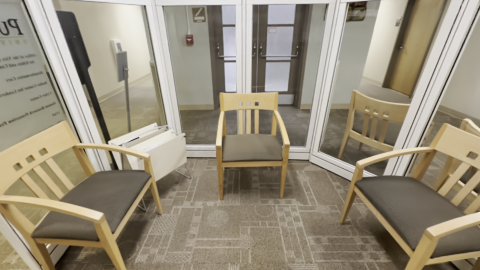}\hspace{1pt}%
        \includegraphics[width=0.195\linewidth]{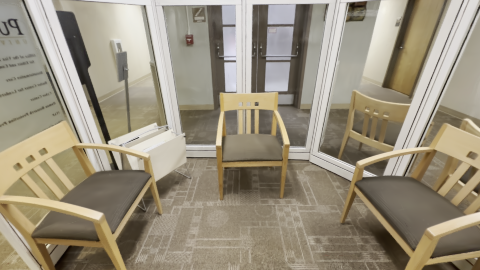}\hspace{1pt}%
        \includegraphics[width=0.195\linewidth]{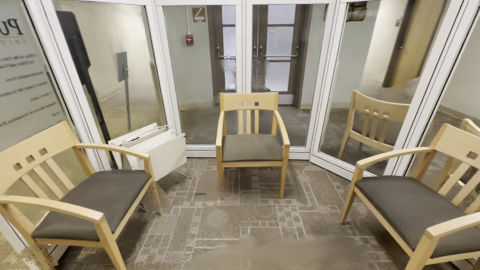}\hspace{1pt}%
        \includegraphics[width=0.195\linewidth]{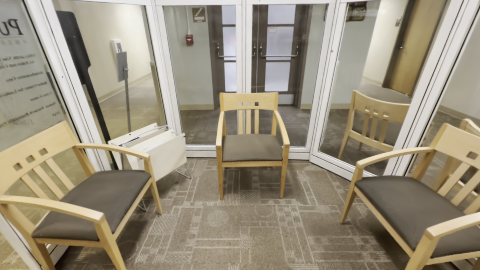}\\[1pt]
        \hspace{0.195\linewidth}\hspace{1pt}%
        \includegraphics[width=0.195\linewidth]{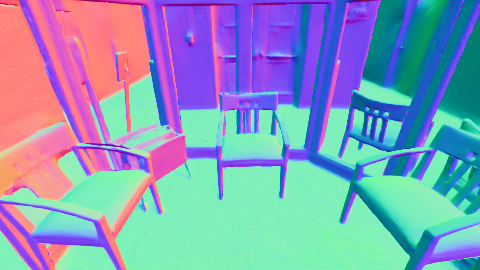}\hspace{1pt}%
        \includegraphics[width=0.195\linewidth]{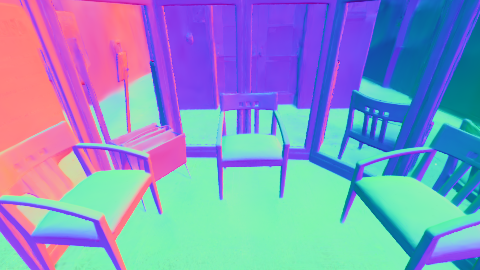}\hspace{1pt}%
        \includegraphics[width=0.195\linewidth]{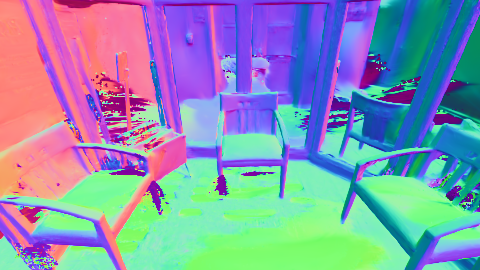}\hspace{1pt}%
        \includegraphics[width=0.195\linewidth]{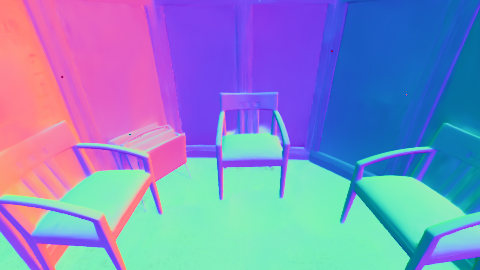}
    \end{subfigure}
    
    \vspace{3pt}
    
    \begin{subfigure}[t]{\linewidth}
        \centering
        \includegraphics[width=0.195\linewidth]{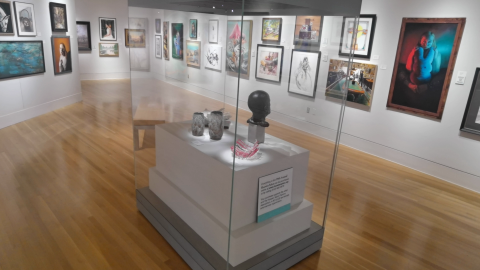}\hspace{1pt}%
        \includegraphics[width=0.195\linewidth]{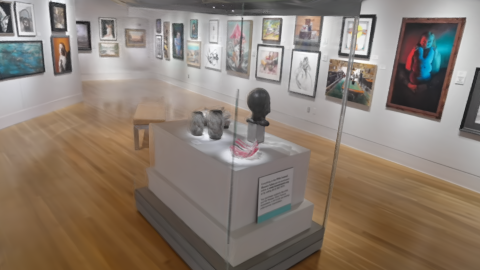}\hspace{1pt}%
        \includegraphics[width=0.195\linewidth]{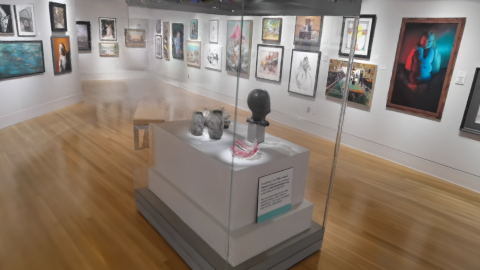}\hspace{1pt}%
        \includegraphics[width=0.195\linewidth]{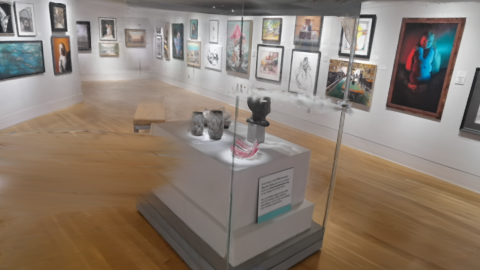}\hspace{1pt}%
        \includegraphics[width=0.195\linewidth]{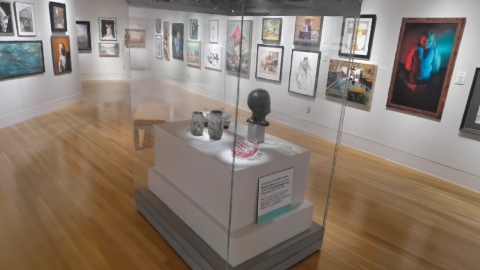}\\[2pt]
        \hspace{0.195\linewidth}\hspace{1pt}%
        \includegraphics[width=0.195\linewidth]{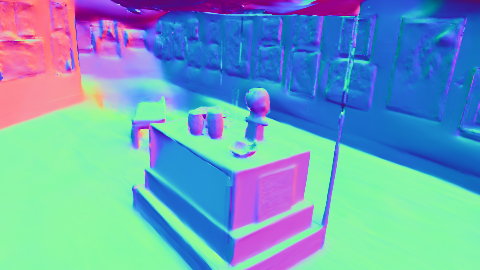}\hspace{1pt}%
        \includegraphics[width=0.195\linewidth]{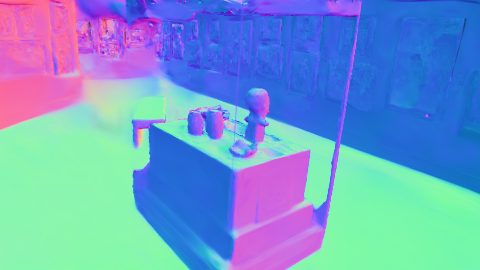}\hspace{1pt}%
        \includegraphics[width=0.195\linewidth]{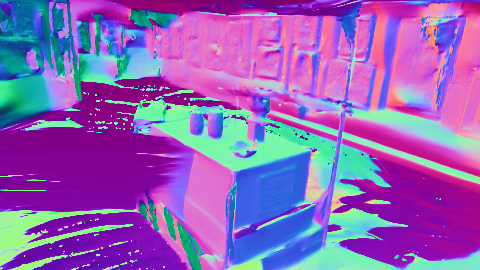}\hspace{1pt}%
        \includegraphics[width=0.195\linewidth]{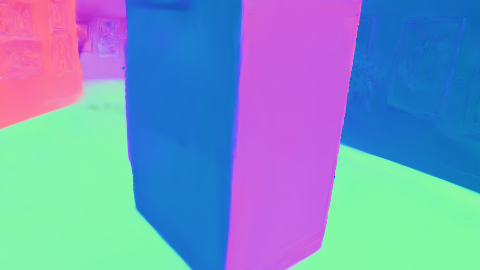}
    \end{subfigure}
    
    \vspace{-3pt}
    
    \caption{
        \textbf{Qualitative comparison on DL3DV-10K dataset.}
        Each column shows results from GT, PGSR~\cite{chen2024pgsr}, EnvGS~\cite{xie2025envgs}, TSGS~\cite{li2025tsgs}, and Ours.
        For each scene, rows correspond to RGB (top) and normal (bottom) maps.
    }
    \label{fig:qual_dl3dv}
\vspace{-7pt}
\end{figure*}

%% file: contents/4_experiments.tex
\section{Experimental Results}
\label{experiments}

\subsection{Implementation Details}
We implement GLINT in PyTorch, integrating the 2DGS rasterizer~\cite{huang20242d} for primary interface rendering and a modified OptiX~\cite{parker2010optix}-based ray tracer adapted from EnvGS~\cite{xie2025envgs} for secondary transmission and reflection queries. All experiments are conducted on a single NVIDIA RTX 4090 GPU.
Following~\cite{kerbl20233d}, we adopt adaptive densification and pruning, augmented with edge-aware normal smoothing and a normal consistency constraint between the rendered normal map and the depth-map gradients, as used in prior works~\cite{huang20242d,li2025tsgs,xie2025envgs}.
For transparency bootstrapping, we set $\tau_d = 0.01$ and $\gamma_a = 0.05$ across all experiments.
Additional implementation details are provided in the Appendix.

\vspace{-4pt}
\subsection{Datasets}
\vspace{-5pt}
We evaluate GLINT on both real-world and synthetic benchmarks to evaluate its performance in reconstructing geometry and appearance in scene-scale transparency.

\vspace{-6pt}
\paragraph{Real-world.}
We leverage DL3DV-10K~\cite{ling2024dl3dv}, a large-scale dataset containing diverse indoor and outdoor scenes with complex material properties, including strong reflection and transmission effects. 
We select a subset of 8 scenes featuring prominent transparent surfaces such as glass partitions, display cases, and windows. 
Since DL3DV-10K lacks ground-truth geometry annotations, the evaluation focuses on photometric quality and qualitative evaluation on geometric reconstruction.

\vspace{-9pt}
\paragraph{Synthetic.}
To enable rigorous quantitative evaluation on transparent geometry, we introduce 3D-FRONT-T, a new synthetic benchmark specifically designed for scene-scale transparency. 
3D-FRONT-T extends 3D-FRONT~\cite{fu20213d} by randomly placing thin transparent elements—such as glass panels and display cases—that enclose or interact with opaque objects, rendered with Blender~\cite{Hess:2010:BFE:1893021}.
The dataset contains 5 scenes with ground-truth depth and normal maps.
Further details are provided in the Appendix.

\subsection{Baselines and Metrics}
\paragraph{Baselines.} 
We compare against representative Gaussian splatting methods that enhance geometric fidelity or model complex non-Lambertian effects. Our baselines include planar-constrained variants (2DGS~\cite{huang20242d}, PGSR~\cite{chen2024pgsr}) and approaches specialized for strong specular reflection (Ref-GS~\cite{zhang2025ref}, EnvGS~\cite{xie2025envgs}). We further include TSGS~\cite{li2025tsgs}, which specifically targets the surface reconstruction of transparent surfaces. We follow the official implementation to reproduce their results. Detailed descriptions of all baselines are provided in the Appendix.

\vspace{-10pt}
\paragraph{Evaluation metrics.}
For rendering quality, we report PSNR, SSIM~\cite{wang2004image}, and LPIPS~\cite{zhang2018unreasonable}.
For geometry reconstruction, we follow Metric3Dv2~\cite{hu2024metric3d} to evaluate depth using absolute relative error (AbsRel), root mean squared error (RMSE), and threshold accuracies $\delta<1.25$.
For normals, we report the mean angular error (MAE) and angular accuracies under $11.25^\circ$, and $22.5^\circ$, representing the percentage of pixels whose estimated normals deviate from the ground truth by less than the specified angle.
For mesh evaluation, we report Chamfer Distance (CD) and F1 score to assess surface reconstruction quality. The original metric for CD is in meters; we report CD in decimeters for readability.

\input{tables/Table_1_NVS}

\vspace{-10pt}
\subsection{Baseline Comparisons}

\paragraph{Quantitative evaluation.}
As shown in Tab.~\ref{tab:geo} and Tab.~\ref{tab:nvs}, GLINT achieves state-of-the-art performance on both real-world (DL3DV-10K~\cite{ling2024dl3dv}) and synthetic (3D-FRONT-T) benchmarks. 
While TSGS~\cite{li2025tsgs} attains reasonable geometric accuracy, its overall rendering quality remains low. In contrast, GLINT delivers both the highest photometric quality (Tab.~\ref{tab:nvs}) and the lowest geometric errors (Tab.~\ref{tab:geo}), consistently outperforming all baselines. 
These results support our core hypothesis that explicitly decomposing radiance into interface, transmission, and reflection components effectively resolves the geometric ambiguities inherent to transparent scenes.

\vspace{-7pt}
\paragraph{Qualitative evaluation.}
Figs.~\ref{fig:qual_synthetic} and~\ref{fig:qual_dl3dv} present qualitative comparisons on synthetic and real-world datasets, demonstrating the superior capability of our approach in reconstructing scene-scale transparency. 
Baseline methods~\cite{chen2024pgsr,xie2025envgs,li2025tsgs} exhibit missing or noisy geometry on glass regions, with corresponding normal and depth maps revealing either incomplete glass surfaces or noisy floating artifacts.
While TSGS~\cite{li2025tsgs} partially alleviates this issue through first-surface transparency modeling with data-driven priors~\cite{ye2024stablenormal,ren2024grounded}, it still fails to recover transmitted radiance, resulting in blurred rendering of objects observed through glass. 
In contrast, GLINT successfully reconstructs well-defined transparent surfaces while simultaneously preserving accurate transmitted appearance and reflection details. 
We further visualize reconstructed meshes obtained via TSDF fusion from the rendered depth maps, as shown in Fig.~\ref{fig:mesh_visualization}. Compared to baseline methods, GLINT produces more coherent glass interfaces and reduces floating artifacts around transparent structures.

\begin{figure}[t]
    \centering
    \vspace{-5pt}
    \setlength{\tabcolsep}{2pt}
    \small
    \begin{tabular}{cc}
        \includegraphics[width=0.44\linewidth]{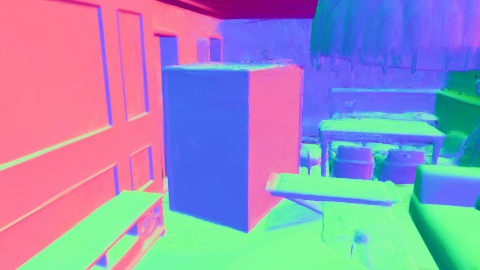} &
        \includegraphics[width=0.44\linewidth]{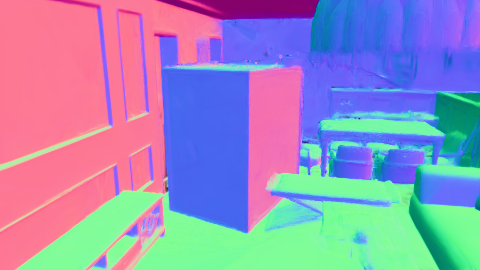} \\[-2pt]
        (a) Ours &
        (b) w/o $\mathcal{L}_{\text{depth}}$ \\[2pt]
        \includegraphics[width=0.44\linewidth]{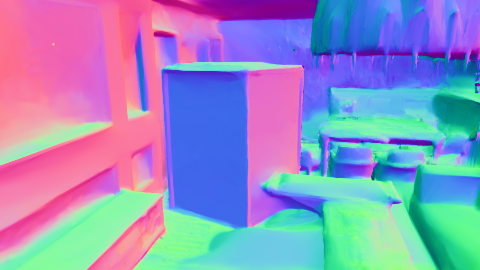} &
        \includegraphics[width=0.44\linewidth]{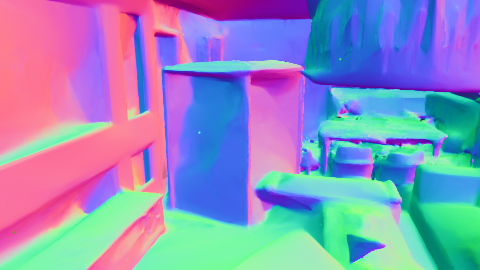} \\[-2pt]
        (c) w/o $\mathcal{L}_{\text{normal}}$ &
        (d) w/o $\mathcal{L}_{\text{geo}}$ \\
    \end{tabular}
    \vspace{-8pt}
    \caption{\textbf{Qualitative Ablation study of G-buffer guidance.}}
    \vspace{-5pt}
    \label{fig:qual_ablation_geo}
\end{figure}

\input{tables/Table_3_ablation}

\subsection{Ablation Studies}
We systematically evaluate each component's contribution in Tab.~\ref{tab:ablation}.
Ablating the transmission component ($\mathcal{G}_{\text{trans}}$) causes the most significant performance drop, as background content incorrectly blends with the interface Gaussians, creating geometric ambiguities that degrade both rendering quality and depth accuracy.
Removing the reflection component ($\mathcal{G}_{\text{refl}}$) moderately reduces rendering fidelity, particularly in specular regions, though geometric metrics remain relatively stable.
The transparency bootstrapping loss ($\mathcal{L}_{\text{trans}}$) serves as a stabilizing regularizer that guides the learning of transparency, yielding more consistent rendering and geometry.
The geometric regularization losses, $\mathcal{L}_{\text{normal}}$ and $\mathcal{L}_{\text{depth}}$, respectively enhance surface orientation and depth accuracy. 
As shown in Fig.~\ref{fig:qual_ablation_geo}, our G-buffer guidance losses effectively regularize the interface geometry.
Additional qualitative ablation results are provided in the Appendix.

%% file: tables/Table_1_NVS.tex
\begin{table}[t!]
\centering
\footnotesize
\setlength{\tabcolsep}{3pt}
\caption{ \textbf{Photometric evaluation on real and synthetic datasets.} GLINT achieves state-of-the-art rendering quality, quantitatively outperforming all baseline methods on both benchmarks.}
\vspace{-5pt}
\begin{tabular}{
  l
  c@{\hspace{4pt}}c@{\hspace{4pt}}c
  c@{\hspace{4pt}}c@{\hspace{4pt}}c
}
\toprule
\multirow{2}{*}{Method} &
\multicolumn{3}{c}{DL3DV-10K~\cite{ling2024dl3dv}} &
\multicolumn{3}{c}{3D-FRONT-T} \\
\cmidrule(lr){2-4} \cmidrule(lr){5-7}
& PSNR$\uparrow$ & SSIM$\uparrow$ & LPIPS$\downarrow$
& PSNR$\uparrow$ & SSIM$\uparrow$ & LPIPS$\downarrow$ \\
\midrule
2DGS~\citep{huang20242d} &
29.18 & 0.91 & 0.13 &
32.26 & \cellthird 0.94 & 0.08 \\

PGSR~\citep{chen2024pgsr} &
\cellthird29.26 & \cellfirst0.92 &  \cellfirst0.11 &
31.96 & \cellsecond 0.95 & \cellthird 0.07 \\

Ref-GS~\citep{zhang2025ref} &
29.11 & \cellfirst0.92 & \cellthird 0.12 &
32.30 & 0.94 & 0.09 \\

EnvGS~\citep{xie2025envgs} &
\cellsecond 29.65 & 0.91 & \cellthird 0.12 &
\cellsecond33.71 & 0.94 & \cellsecond 0.07 \\

TSGS~\citep{li2025tsgs} &
25.94 & 0.85 & 0.19 &
28.80 & 0.87 & 0.14 \\

\textbf{GLINT (Ours)} &
\cellfirst 30.21 & \cellfirst 0.92 & \cellfirst 0.11 &
\cellfirst 34.50 & \cellfirst 0.96 & \cellfirst 0.05 \\

\bottomrule
\end{tabular}
\vspace{-5pt}
\label{tab:nvs}
\end{table}

%% file: tables/Table_3_ablation.tex
\begin{table}[t!]
\centering
\small
\renewcommand{\arraystretch}{1.0}
\setlength{\tabcolsep}{3pt}
\caption{\textbf{Ablation studies on our method.} We report PSNR, SSIM, LPIPS, Normal MAE, and Depth AbsRel.}
\vspace{-3pt}
\begin{tabular}{lccccc}
\toprule
Method & PSNR $\uparrow$ & SSIM $\uparrow$ & LPIPS $\downarrow$ & MAE $\downarrow$ & AbsRel $\downarrow$ \\
\midrule
\textit{w/o} $\mathcal{G}_{\text{trans}}$  & 32.26 & 0.93 & 0.085 & 8.11 & 0.038 \\
\textit{w/o} $\mathcal{G}_{\text{refl}}$   & 32.70 & 0.94 & 0.067 & 8.78 & 0.038 \\
\textit{w/o} $\mathcal{L}_{\text{trans}}$  & 33.57 & 0.94 & 0.066 & 8.07 & 0.037 \\
\textit{w/o} $\mathcal{L}_{\text{normal}}$ & 33.92 & 0.95 & 0.060 & 12.21 & 0.043 \\
\textit{w/o} $\mathcal{L}_{\text{depth}}$  & 34.05 & 0.95 & 0.058 & 8.54 & 0.061 \\
\textit{w/o} $\mathcal{L}_{\text{geo}}$    & 33.62 & 0.95 & 0.060 & 24.69 & 0.126 \\
\midrule
Full model & 34.50 & 0.96 & 0.048 & 7.96 & 0.035 \\
\bottomrule
\end{tabular}
\vspace{-11pt}
\label{tab:ablation}
\end{table}

%% file: contents/5_discussion.tex
\vspace{-3pt}
\section{Conclusion and Discussion}
\label{sec:conclusion}
In this study, we introduced GLINT, the first framework to reconstruct scene-scale transparency through an explicit decomposition of interface, transmission, and reflection components. This formulation enables coherent geometry and appearance modeling, effectively disentangling transparent regions. Extensive experiments show that GLINT achieves state-of-the-art performance across both appearance and geometry metrics, enabling faithful reconstruction of complex transparent structures.

\noindent\textbf{Limitations.}
While our rendering pipeline focuses on first-order radiance interactions, extending it to handle multi-bounce phenomena including nested transparency remains an important direction for future work.

%% file: contents/X_suppl.tex
\clearpage
\setcounter{page}{1}
\maketitlesupplementary

\appendix

This supplementary material provides additional details on (i) the proposed synthetic 3D-Front-T dataset (Sec.~\ref{sec:supl_dataset}), (ii) the baseline methods used in our experiments (Sec.~\ref{sec:supl_baselines}), and (iii) implementation details (Sec.~\ref{sec:supl_impl}). In addition, we include qualitative results on ablation studies (Sec.~\ref{sec:supl_qual_ablation}), comparative analysis with baselines (Sec.~\ref{sec:supl_extended_analysis}), discussion on foundation models (Sec.~\ref{sec:supl_discussion_on_foundation}), detailed discussion on limitations and future work (Sec.~\ref{sec:supl_limitations_and_future_work}), and additional qualitative results on both real and synthetic datasets (Sec.~\ref{sec:supl_additional_qual}).

\section{3D-FRONT-T Dataset}
\label{sec:supl_dataset}
To enable rigorous quantitative evaluation of scene-scale transparency reconstruction, we introduce 3D-FRONT-T, a new synthetic benchmark for scene-scale transparency reconstruction. We construct our dataset built upon the 3D-FRONT (3D Furnished rooms with
layouts and semantics) dataset~\cite{fu20213d}.
While existing real-world datasets contain transparent scenes~\cite{ling2024dl3dv}, they lack ground-truth geometry annotations for quantitative evaluation on geometry reconstruction.
Our 3D-FRONT-T addresses this by providing depth and normal ground truth alongside RGB renderings of transparent scenes.

\subsection{Dataset Construction}
We collect 5 indoor scenes from the 3D-FRONT~\cite{fu20213d} datasets with diverse configurations. For each scene, we first identify the largest room by floor area and retain only the objects within that room, including walls, ceilings, doors, windows, and furniture placed on the floor.
Then, we texture the floor, wall and ceiling in the scene with the diverse random materials following~\cite{nie2023learning}. 

On top of this setup, we create and put a glass display container with a black metal frame, enclosing opaque objects. The container consists of six thin glass panels with a supporting frame structure. The container size is randomly determined to create different sizes of transparency regions. 

To achieve realistic placement, we utilize Blender’s physics-based simulation~\cite{Hess:2010:BFE:1893021} to settle objects into physically plausible configurations. The glass container is initialized at a random position above the floor and released using rigid-body dynamics, allowing it to naturally come to rest on the floor or on top of existing furniture. This process ensures physically consistent placement while generating diverse occlusion patterns with the surrounding scene geometry.
 
All scenes are rendered using the Blender Cycles path tracer with 4096 samples per pixel and a maximum of 200 light bounces across all channels (diffuse, glossy, and transmission) to accurately capture complex light interactions through transparent surfaces. For our experiments, we render all images at a resolution of 960×540.
For each scene, we generate camera trajectories which samples continuous viewpoints that ensure visibility of key objects including the transparent container. The example images of the dataset are shown in Fig.~\ref{fig:supl_front_t_example}.

\input{figures/supl/supl_3d_front_example}

\subsection{Ground Truth Annotations}

For each rendered viewpoint, we provide a complete set of physically accurate ground-truth annotations tailored for evaluating scene-scale transparency reconstruction:

\noindent
\textbf{RGB Images.}
High-fidelity renderings produced via multi-bounce path tracing in Blender Cycles. These images capture intricate light behavior including interreflections, refractions, and transmission through transparent materials, serving as a challenging benchmark for appearance modeling.

\noindent
\textbf{Depth Maps.}
Metric depth is extracted directly from Blender’s rendering pipeline, ensuring physically consistent geometry even in regions partially occluded or viewed through transparent media. This enables rigorous evaluation of depth recovery in transparent scenes, a regime underexplored in existing benchmarks.

\noindent
\textbf{Normal Maps.}
Per-pixel surface normals are rendered for all visible surfaces. These provide robust supervision signals for assessing fine-grained geometric accuracy, independent of texture or lighting cues.

\noindent
Taken together, these annotations establish a comprehensive benchmark for transparent-scene reconstruction. To encourage reproducibility and extensibility, we will release our full dataset-generation pipeline, implemented on top of BlenderProc~\cite{Denninger2023}, enabling automatic synthesis of large-scale, diverse scenes with configurable and physically plausible object placement.

\input{figures/supl/supl_qual_ablation}
\section{Overview of Baseline Methods}
\label{sec:supl_baselines}

We compare GLINT with representative Gaussian-splatting–based approaches that cover planar-constrained geometry modeling, reflective radiance modeling, and transparent-surface reconstruction. Below, we briefly summarize each baseline to clarify their modeling assumptions and limitations in the context of scene-scale transparency.

\noindent
\textbf{2DGS}~\cite{huang20242d} adopts 2D Gaussians to improve geometric accuracy over volumetric 3DGS. While effective for opaque surfaces, its monolithic $\alpha$-compositing cannot separate interface geometry from secondary effects, causing transparent surfaces to be ignored or entangled into a single depth layer.

\noindent
\textbf{PGSR}~\cite{chen2024pgsr} extends planar-aligned Gaussian primitives with additional geometric constraints. Despite achieving sharper surface reconstruction, it shares the same opacity-based rendering pipeline and therefore struggles with multi-depth radiance, often collapsing glass regions into the background.

\noindent
\textbf{Ref-GS}~\cite{zhang2025ref} models view-dependent appearance through directional factorization, enabling more expressive specular materials. However, its formulation implicitly assumes that all non-Lambertian behavior arises from surface reflection, without accounting for light transmission. As a result, transparent materials whose appearance is governed by both interface reflectance and background transmission are incorrectly treated as purely reflective surfaces, producing biased material estimates and degraded geometric cues in regions where transmitted radiance is dominant.

\noindent
\textbf{EnvGS}~\cite{xie2025envgs} models environment radiance using a dedicated set of Gaussians and performs ray-traced reflection queries, supported by a monocular normal prior to stabilize geometry estimation from limited viewpoints. While highly effective for opaque materials, it lacks any mechanism to account for light interactions through transparent surfaces, leading background content seen through glass to be incorrectly attributed to reflections.

\noindent
\textbf{TSGS}~\cite{li2025tsgs} targets transparent objects using first-surface rasterization combined with monocular normal~\cite{ye2024stablenormal}, delighting~\cite{ye2024stablenormal}, and segmentation priors~\cite{ren2024grounded}. While effective for thin, object-centric transparency, its formulation assumes a single transparent interface and does not explicitly model transmitted radiance. Consequently, scenes containing multiple depth layers (e.g., glass–background–interior structures) often exhibit blurred transmission and incomplete geometry reconstruction. The segmentation module also produces noisy or missing transparency masks, an issue we further analyze in the next section.

Overall, existing baselines either (i) emphasize geometric fidelity, (ii) specialize in reflective appearance modeling, or (iii) operate under object-centric transparency assumptions. None provide a unified framework capable of addressing the inherently ill-posed nature of scene-scale transparency, where accurate reconstruction requires jointly modeling interface geometry, background transmission, and reflection. Our study bridges this gap by introducing a decomposed Gaussian representation and transparency-aware radiance transport, which together provide a more physically consistent formulation for scene-scale transparency.

\section{Additional Implementation Details}
\label{sec:supl_impl}

In this section, we provide comprehensive implementation details for reproducibility, including initialization process, multi-stage optimization, and the specific loss formulations designed to ensure physical plausibility and geometric consistency. Our code will be made publicly available. 

\paragraph{Initialization.} 
To establish a reliable geometric foundation, we initialize the interface Gaussian ($\mathcal{G}_{\text{intr}}$) and transmission Gaussian ($\mathcal{G}_{\text{trans}}$) primitives using the sparse point cloud derived from Structure-from-Motion (SfM)~\cite{fu2024colmap}. Meanwhile, we follow EnvGS~\cite{xie2025envgs} for initializing reflectance component ($\mathcal{G}_{\text{refl}}$), where we partition the scene into $N^3$ sub-grids by partitioning the bounding box. Then we randomly sample $K$ primitives within the grid where we set $N=32$ and $K=5$.

\paragraph{Optimization Schedule.}
To ensure stable training, we adopt a multi-stage optimization strategy that progressively recovers scene geometry before refining complex radiance effects. We begin with a 5k-iteration warm-up stage in which only the interface component is optimized, allowing the primary surface geometry to converge. Afterward, the transmission and reflection components are introduced for joint optimization. Finally, between 40k and 60k iterations, we freeze the interface Gaussians and update only the transmission and reflection Gaussians to refine their radiance behavior without destabilizing the established geometry.

\paragraph{Regularization.}
For geometric regularization, we adopt a scale-and-shift--invariant depth loss together with a normal consistency loss, following the formulation of MonoSDF~\cite{yu2022monosdf}. 
Let $z$ denote the rendered depth from the interface component and $\hat{z}$ the monocular depth prior predicted by the encoder~\cite{liang2025diffusion}. 
We align $z$ and $\hat{z}$ by solving for the optimal scale $w$ and shift $q$ in closed form, and define the depth loss as:
\begin{equation}
\mathcal{L}_{\mathrm{depth}}
=
\frac{1}{N}
\sum_{i=1}^{N}
\left(
    w\, z_i + q - \hat{z}_i
\right)^2,
\label{eq:si_depth}
\end{equation}
where $N$ denotes the number of pixels in the image.
This formulation removes the global scale ambiguity of monocular predictions while preserving their relative depth structure.
The scale--shift parameters $(w, q)$ are recomputed for each image using the closed-form least-squares solution~\cite{yu2022monosdf}.

For the normal loss, we follow the thresholded normal supervision strategy introduced in TSGS~\cite{li2025tsgs} to mitigate the influence of noisy monocular priors. 
Given the rendered normal $\mathbf{n}$ and the normal prior from~\cite{liang2025diffusion} $\hat{\mathbf{n}}$, we apply a cosine-similarity mask from 10k iterations:
\[
\mathbf{M}_{\text{prior}}(u)
=
\big[
\langle \mathbf{n}(u), \hat{\mathbf{n}}(u) \rangle \ge \tau_{\mathrm{n}}
\big],
\]
and compute the masked normal loss as:
\[
\mathcal{L}_{\text{normal}}
=
\sum_{u}
\mathbf{M}(u)
\big(
1 - 
\langle
\mathbf{n}(u), \hat{\mathbf{n}}(u)
\rangle
\big).
\]
We set $\tau_{\mathrm{n}} = 0.3$ for all experiments.

\section{Qualitative Ablation Studies}
\label{sec:supl_qual_ablation}

\input{figures/supl/supl_qual_ablation_geo}
\input{figures/supl/supl_qual_trans}

In this section, we present additional qualitative ablation results to further elucidate the distinct roles of each component within the GLINT framework. 

First, we examine the impact of our decomposed representation in Fig.~\ref{supl:qual_ablation}. Removing the transmission branch ($\mathcal{G}_{\text{trans}}$) leads to significant entanglement between the interface and background content, resulting in geometric inconsistencies and a washed-out appearance in regions observed behind glass. Excluding the reflection branch ($\mathcal{G}_{\text{refl}}$) diminishes specular cues, leading to a loss of realistic surface gloss, particularly on glass or highly reflective surfaces.

Next, we visualize the ablation of geometry regularization losses in Fig.~\ref{fig:supl_qual_ablation_geo}. While our framework remains relatively robust, removing specific losses introduces characteristic degradations that highlight their individual contributions. Excluding the depth loss ($\mathcal{L}_{\text{depth}}$) causes inaccuracies in interface geometry placement, manifesting as deviations in absolute depth, especially for transparent surfaces. Removing the normal loss ($\mathcal{L}_{\text{normal}}$) results in less consistent surface orientation, which appears as locally unstable shading and noisy normal transitions. When all geometric priors are removed ($\mathcal{L}_{\text{geo}}$), these artifacts accumulate, leading to noticeable distortions in depth and normal maps. This indicates that the geometric constraints operate complementarily to maintain structural fidelity.

Finally, we analyze the effect of the transparency bootstrapping loss ($\mathcal{L}_{\text{trans}}$) in Fig.~\ref{fig:trans_ablation}. Omitting this loss yields noisier and spatially inconsistent transparency estimates.
This is because the loss utilizes 3D geometric cues derived from depth separation to guide the optimization toward sharp and physically consistent transparency boundaries.

\input{figures/supl/radiance_decomp_compare}
\input{figures/supl/envgswdiffren}
\input{figures/supl/supl_sam_comparison}

\section{Comparative Analysis with Baselines}
\label{sec:supl_extended_analysis}

In this section, we provide a comprehensive comparative analysis against baseline methods. We focus on three key aspects: the interpretability of radiance decomposition, the accuracy of transparency localization, and a quantitative analysis of computational costs versus reconstruction quality.

\paragraph{Radiance decomposition.}
Fig.~\ref{fig:radiance_decomp} presents an extended comparison of radiance decomposition between our method and EnvGS~\cite{xie2025envgs}. 

For our method (Fig.~\ref{fig:radiance_decomp}(a)), the decomposed Gaussian representation enables a clear and physically grounded partitioning of radiance. The interface component isolates the surface base color, while the reflection component captures specular highlights. Crucially, the transmission component reconstructs the background scene visible specifically through transparent surfaces, revealing the correct spatial structure behind the glass. These components form a coherent explanation of the observed radiance, demonstrating that our explicit decomposition naturally disentangles overlapping radiance sources in transparent regions.

In contrast, EnvGS (Fig.~\ref{fig:radiance_decomp}(b)) focuses solely on modeling view-dependent reflections via environment Gaussians and lacks a dedicated transmission component. Consequently, radiance originating from behind transparent surfaces is incorrectly entangled within its diffuse or reflection modeling, leading to mixed or incomplete visual explanations. Furthermore, its reflection strength is modulated by a single scalar weight $s$, lacking the physically accurate Fresnel-driven angular dependence inherent to our formulation.

To further validate the necessity of our decomposition, we conducted an experiment training EnvGS~\cite{xie2025envgs} with the geometric normal priors from DiffusionRenderer~\cite{liang2025diffusion}, identical to our setup. As illustrated in Fig.~\ref{fig:envgs_diffren_compare}, enforcing geometric consistency on EnvGS paradoxically leads to corrupted appearance rendering. This degradation occurs because EnvGS lacks a dedicated transmission component and fundamentally conflates transmitted and reflected radiance into a single surface interaction. Typically, EnvGS implicitly minimizes photometric error by distorting the geometry to effectively baking background textures onto incorrect depths. However, when the surface geometry is enforced via stronger priors, this compensatory mechanism is blocked. Consequently, the model is forced to approximate the superposition of multiple radiance layers onto a single interface, leading to an averaging effect that manifests as severe blurring.

\paragraph{Transparency Map Comparison.}
We compare our learned transparency maps with those generated by Grounded-SAM-2 (G-SAM2)~\cite{ren2024grounded}, a state-of-the-art open-world segmentation model. As illustrated in Fig.~\ref{fig:trans_mask_two_scenes}, although G-SAM2 can roughly localize transparent objects, it frequently yields spatially inconsistent or fragmented masks. Common failure cases include missing sections of glass cabinets or exhibiting severe flickering across viewpoints, even when utilizing the tracking mode. These limitations arise because G-SAM2 relies on 2D image features, which are inherently ambiguous for transparent surfaces that mix reflection and transmission, lacking a unified understanding of 3D geometry.

In contrast, GLINT bootstraps transparency localization by explicitly leveraging 3D geometric cues—specifically, the depth discrepancy ($\Delta z$) between the interface and transmission components that emerges during optimization and the diffuse-albedo prior from~\cite{liang2025diffusion}. This allows our method to generate spatially coherent and boundary-sharp transparency maps that accurately align with the physical extent of the glass.

\paragraph{Computational Costs.}
We report a runtime comparison with EnvGS and
TSGS on the synthetic 3D-FRONT-T dataset (downscale $\times2$),
with results summarized in Tab.~B.
TSGS achieves the highest speed due to rasterization-only rendering,
while EnvGS adopts a hybrid formulation limited to reflection.
Our method introduces additional overhead from explicit multi-component
optimization and transmission handling, resulting in lower throughput.
However, this design choice directly supports transparent geometry modeling
and leads to improved reconstruction quality, reflecting a deliberate and
practical trade-off between efficiency and capability.

\begin{table}[h]
    \centering
    \caption{Comparison of computational costs and reconstruction quality on the 3D-FRONT-T dataset.}
    \label{tab:supl_costs}
    \resizebox{\linewidth}{!}{
    \begin{tabular}{l|cc|ccc}
    \toprule
    Method & FPS ($\uparrow$) & Training Time ($\downarrow$) & 
    PSNR ($\uparrow$) & MAE ($\downarrow$) & AbsRel ($\downarrow$) \\
    \midrule
    TSGS~\cite{li2025tsgs} & \textbf{159} & \textbf{$\sim$40 mins} & 28.80 & 9.89 & 0.08 \\
    EnvGS~\cite{xie2025envgs} & 80 & $\sim$1 h & 33.71 & 14.37 & 0.13 \\
    \textbf{Ours} & 51 & $\sim$2.5 h & \textbf{34.50} & \textbf{7.96} & \textbf{0.04} \\
    \bottomrule
    \end{tabular}
    }
\end{table}

\section{Discussion on Foundation Models}
\label{sec:supl_discussion_on_foundation}

In this section, we discuss the motivation behind integrating foundation models into the GLINT framework and analyze the advantages of video-based priors compared to conventional image-based approaches in the context of transparent scene reconstruction.

In our implementation, we utilize the inverse-rendering encoder of the video relighting model, DiffusionRenderer~\cite{liang2025diffusion}, to obtain geometric (depth, normal) and material (diffuse-albedo) priors.
A key motivation for choosing a video-based foundation model over monocular estimators is its multi-view consistency. Since the model leverages the architecture of Stable Video Diffusion~\cite{blattmann2023stable}, it processes multiple frames in a single feed-forward pass, producing geometric cues that are coherent across viewpoints. This is particularly crucial for transparent surfaces, where per-frame estimation often flickers or yields inconsistent depth due to view-dependent reflections. 
To the best of our knowledge, GLINT is the first approach to adopt video relighting priors for 3D Gaussian splatting reconstruction, demonstrating that material-aware priors effectively aid in distinguishing between interface and transmitted radiance.

\paragraph{Comparison with Image-based Priors.}
Recent approaches~\cite{xie2025envgs,li2025tsgs,tong2025gs} typically combine various image-based foundation models, such as monocular depth~\cite{hu2024metric3d,Bochkovskii2024:arxiv}, normal estimators~\cite{ye2024stablenormal,ke2023repurposing}, and segmentation modules (e.g., Grounded-SAM~\cite{kirillov2023segment,ren2024grounded}). 
For instance, TSGS~\cite{li2025tsgs} relies on Grounded-SAM to mask transparent objects and uses StableDelight and StableNormal~\cite{ye2024stablenormal} to make the texture of the transaprent surface dinstinctive. While empirically beneficial for object-centric scenes with clear boundaries, this mask strategy often struggles with scene-scale transparency—such as large glass facades or windows—where semantic boundaries are ambiguous and binary segmentation fails to capture the continuous transition of radiance as in Fig.~\ref{fig:trans_mask_two_scenes}.

\section{Limitations and Future Work}
\label{sec:supl_limitations_and_future_work}

While GLINT achieves state-of-the-art performance in reconstructing scene-scale transparent geometry and appearance, there are still room for improvement in various perspectives. In this section, we detail the discussion on the current boundary conditions of our framework and suggest potential directions to address them.

\vspace{1mm}
\noindent\textbf{Decomposition ambiguity under sparse observations.}
Our method implicitly relies on multi-view consistency to disentangle the intertwined radiance contributions from interface reflection and background transmission. Consequently, in scenarios with sparse viewpoints or limited parallax, where a surface is observed from a stationary angle, the problem becomes inherently ill-posed. In such cases, the optimization may struggle to uniquely assign radiance to either the reflection ($\mathcal{G}_\text{refl}$) or transmission ($\mathcal{G}_\text{trans}$) component. We believe that incorporating high-level semantic understanding could resolve this ambiguity. Future work could leverage vision-language models (VLMs) or unprojecting semantic features to enforce physically and semantically plausible decomposition, ensuring that regions identified as glass (e.g., windows vs. mirrors) adhere to their expected optical behaviors even under constrained observation.

\vspace{1mm}
\noindent\textbf{Recursive light transport.}
To maintain rendering efficiency and optimization stability, our current radiance transport formulation focuses on primary transmission and reflection events (i.e., up to first-order interactions at the interface). While this approximation is sufficient for most architectural glass and display cases, it may not fully capture complex multi-bounce phenomena found in nested transparent structures, such as a glass vase inside a glass cabinet or a mirror room. Extending our hybrid rendering pipeline to support recursive ray-tracing or multi-pass rendering with physically-based rendering formulation would allow for the simulation of higher-order approximation of light transport, albeit at the cost of increased computational overhead.

\section{Additional Qualitative Results}
\label{sec:supl_additional_qual}

We present additional qualitative examples to complement the results in the main paper.
Figures~\ref{fig:supl_additional_qual_3d_front_t} and~\ref{fig:supl_additional_qual_dl3dv} provide further visualizations on representative scenes from both the 3D-FRONT-T benchmark and the real-world DL3DV-10K dataset~\cite{ling2024dl3dv}.

We also include the video results for the visualization of the continuous frames. We kindly refer the readers to the attached supplementary videos.

\input{figures/supl/supl_qual_3d_front_t}
\input{figures/supl/supl_qual_dl3dv}

%% file: figures/supl/supl_3d_front_example.tex
\begin{figure}[t!]
    \centering

    \begin{subfigure}{0.485\linewidth}
        \centering
        \includegraphics[width=\linewidth]{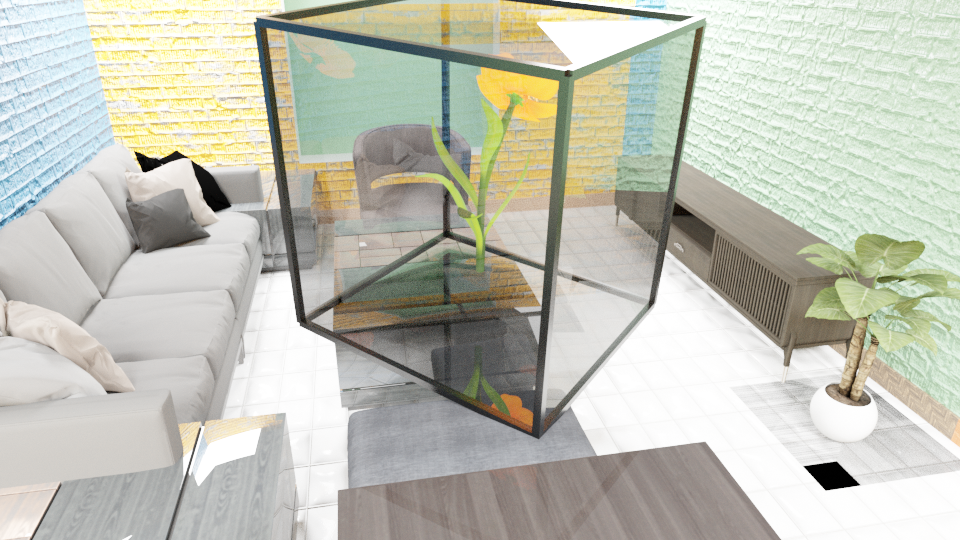}
    \end{subfigure}
    \hspace{2pt}
    \begin{subfigure}{0.485\linewidth}
        \centering
        \includegraphics[width=\linewidth]{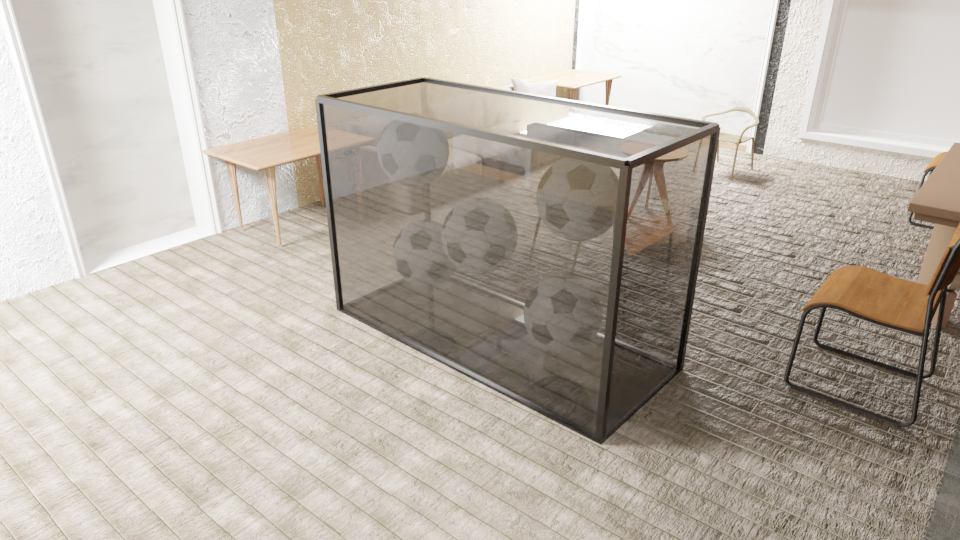}
    \end{subfigure}

    {(a) RGB}

    \begin{subfigure}{0.485\linewidth}
        \centering
        \includegraphics[width=\linewidth]{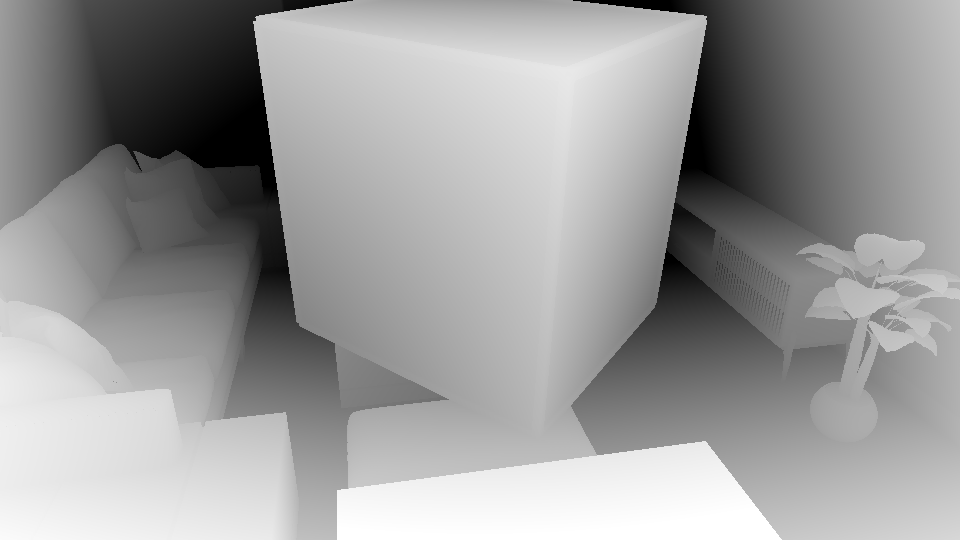}
    \end{subfigure}
    \hspace{2pt}
    \begin{subfigure}{0.485\linewidth}
        \centering
        \includegraphics[width=\linewidth]{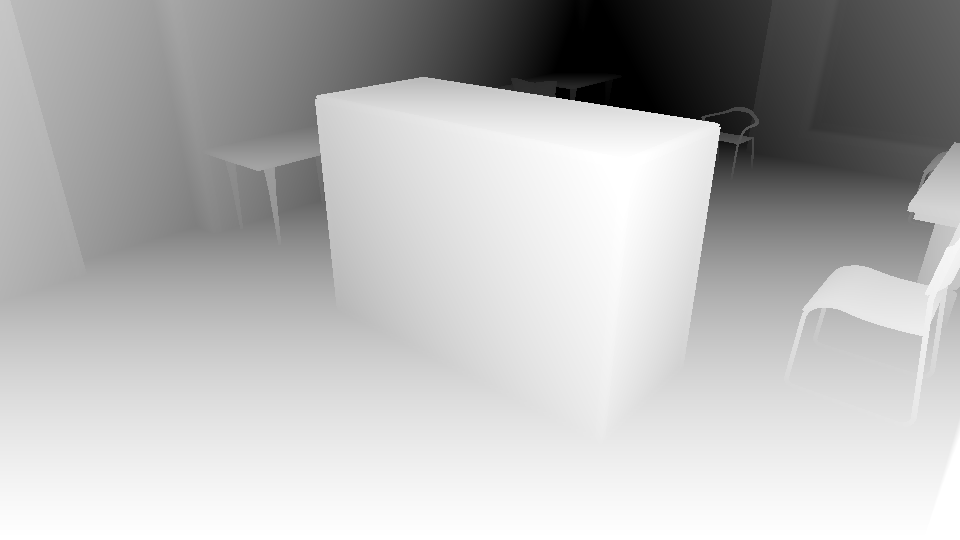}
    \end{subfigure}

    {(b) GT Depth}

    \begin{subfigure}{0.485\linewidth}
        \centering
        \includegraphics[width=\linewidth]{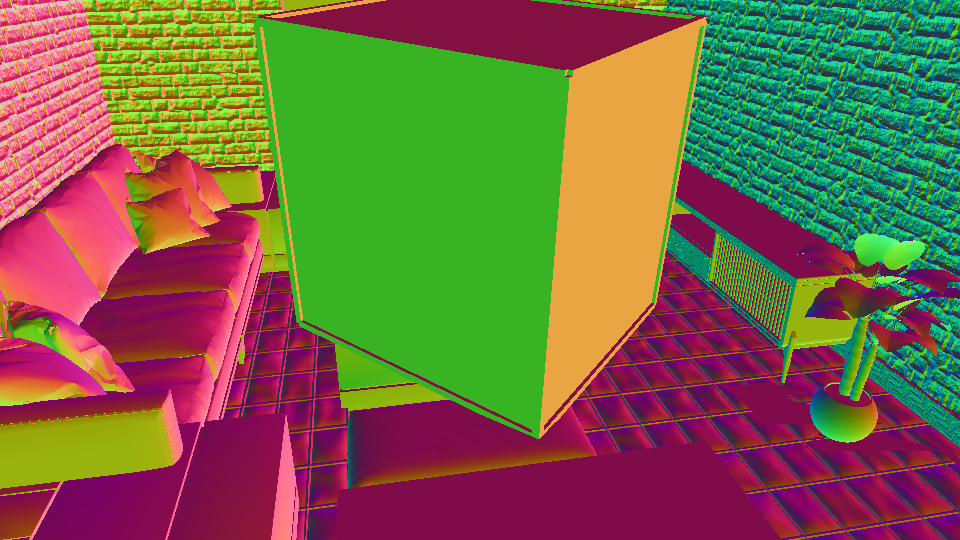}
    \end{subfigure}
    \hspace{2pt}
    \begin{subfigure}{0.485\linewidth}
        \centering
        \includegraphics[width=\linewidth]{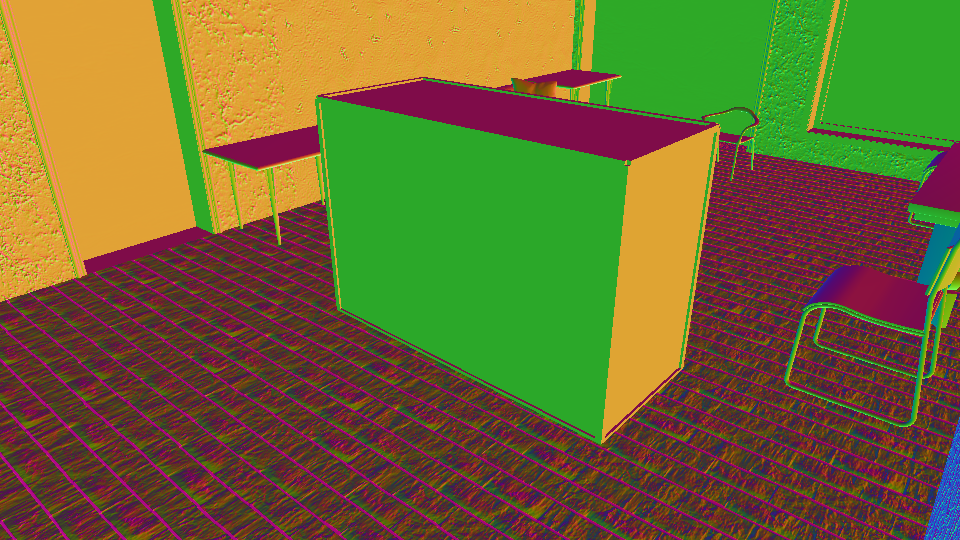}
    \end{subfigure}

    {(c) GT Normal}

    \caption{\textbf{3D-FRONT-T dataset.}
    RGB, depth, and normal ground truth examples. As can be seen in the transparent surface (glass), the outgoing radiance for each pixel is entanglement of radiance from interface surface and the transmitted radiance.}
    \label{fig:supl_front_t_example}
\end{figure}

%% file: figures/supl/supl_qual_ablation.tex
\begin{figure*}[ht!]
    \centering
    \setlength{\tabcolsep}{2pt}

    \renewcommand{\arraystretch}{0.9}
    \begin{tabular}{cccc}
        \makebox[0.24\linewidth][c]{\textrm{GT}} &
        \makebox[0.24\linewidth][c]{\textrm{Full Model}} &
        \makebox[0.24\linewidth][c]{\textrm{\textit{w/o} $\mathcal{G}_{\text{trans}}$}} &
        \makebox[0.24\linewidth][c]{\textrm{\textit{w/o} $\mathcal{G}_{\text{refl}}$}}
    \end{tabular}

    \vspace{2pt}

    \includegraphics[width=0.24\linewidth]{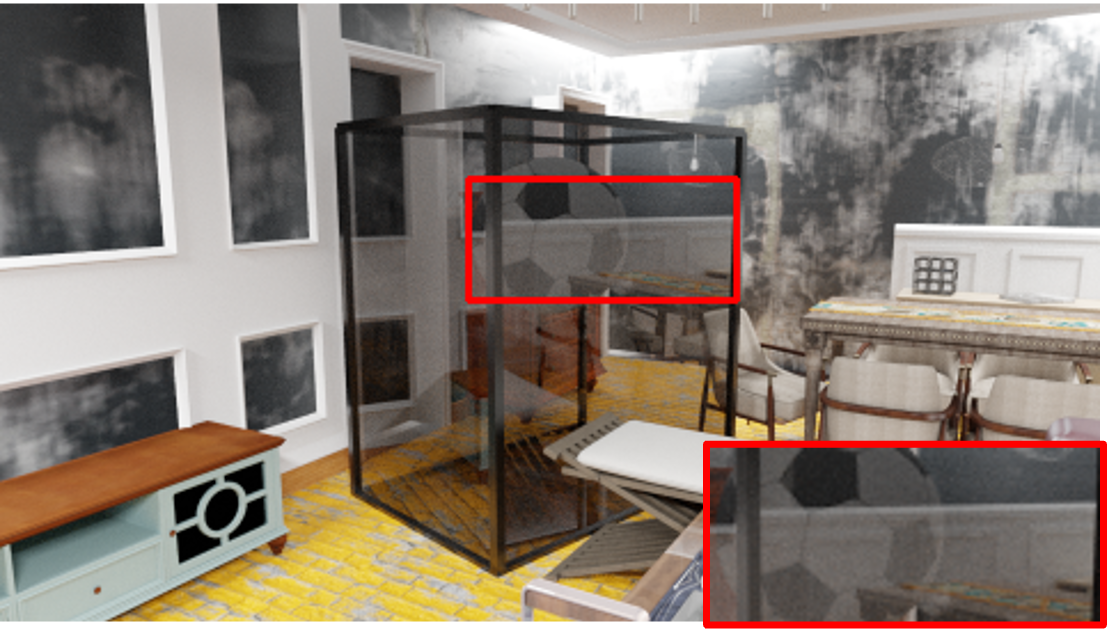}%
    \hspace{1pt}%
    \includegraphics[width=0.24\linewidth]{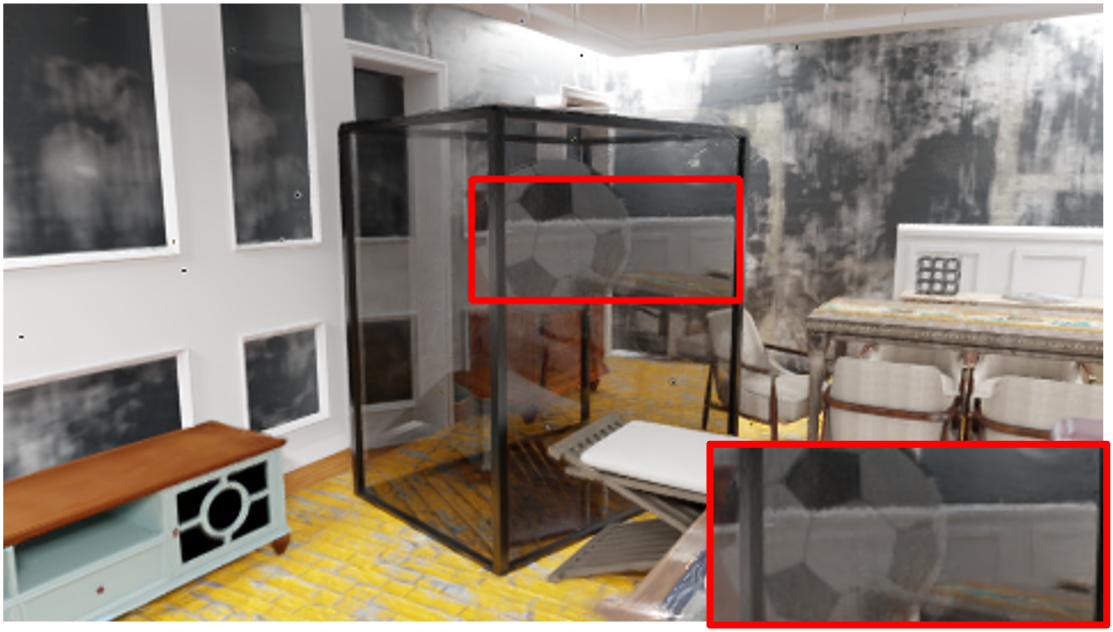}%
    \hspace{1pt}%
    \includegraphics[width=0.24\linewidth]{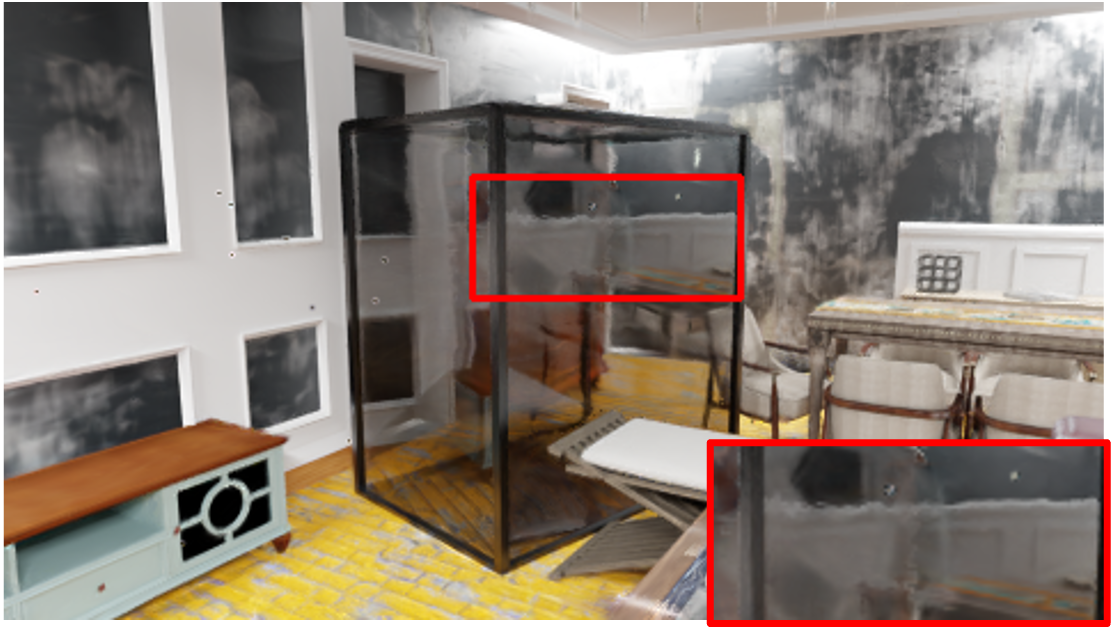}%
    \hspace{1pt}%
    \includegraphics[width=0.24\linewidth]{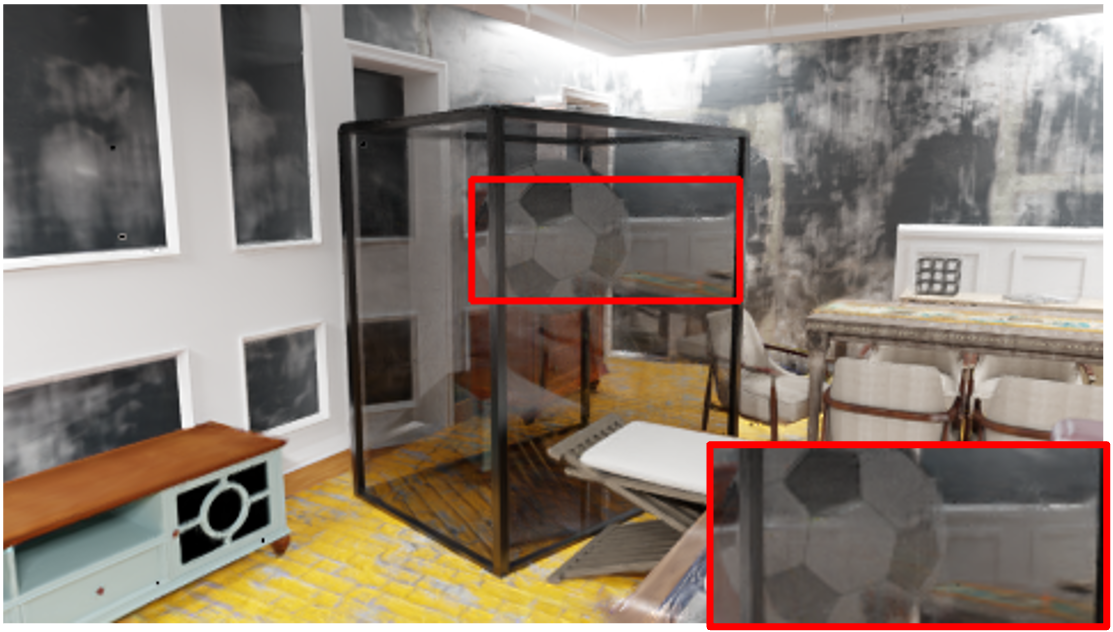}

    \caption{
        \textbf{Qualitative ablation study on representation components.}  
        Visual comparison between the full model and ablated variants.
    }
    \label{supl:qual_ablation}
    \vspace{-6pt}
\end{figure*}

%% file: figures/supl/supl_qual_ablation_geo.tex
\begin{figure*}[t]
    \centering
    \setlength{\tabcolsep}{2pt}

    \begin{tabular}{ccc}
    \end{tabular}

    \vspace{2pt}

    \begin{subfigure}{0.32\linewidth}
        \includegraphics[width=\linewidth]{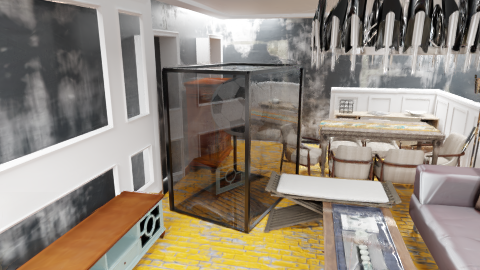}
    \end{subfigure}
    \begin{subfigure}{0.32\linewidth}
        \includegraphics[width=\linewidth]{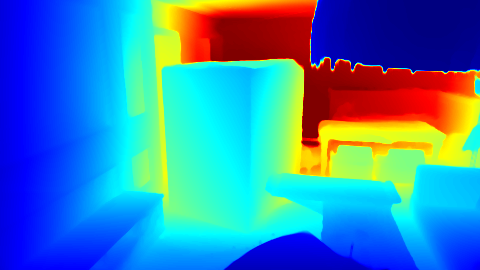}
    \end{subfigure}
    \begin{subfigure}{0.32\linewidth}
        \includegraphics[width=\linewidth]{fig_sources/supl/qual_ablation_geo/ours_scene_4_normal_0016.png}
    \end{subfigure}

    (a) Ours
    \vspace{2pt}

    \begin{subfigure}{0.32\linewidth}
        \includegraphics[width=\linewidth]{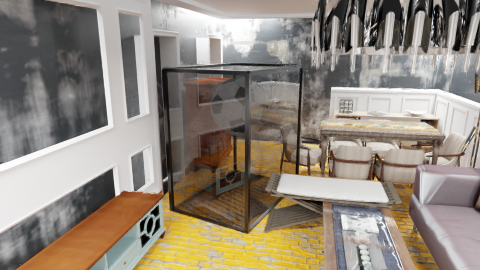}
    \end{subfigure}
    \begin{subfigure}{0.32\linewidth}
        \includegraphics[width=\linewidth]{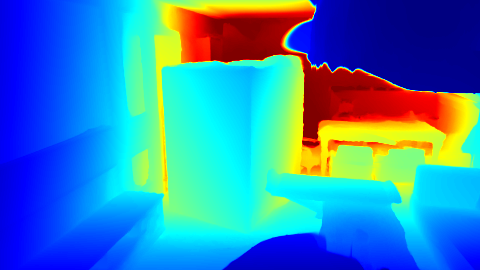}
    \end{subfigure}
    \begin{subfigure}{0.32\linewidth}
        \includegraphics[width=\linewidth]{fig_sources/supl/qual_ablation_geo/wodepth_scene_4_normal_0016.png}
    \end{subfigure}

    (b) w/o $\mathcal{L}_{\text{depth}}$
    \vspace{2pt}

    \begin{subfigure}{0.32\linewidth}
        \includegraphics[width=\linewidth]{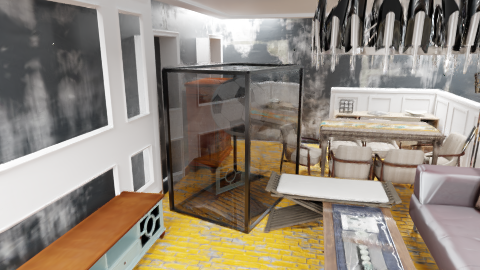}
    \end{subfigure}
    \begin{subfigure}{0.32\linewidth}
        \includegraphics[width=\linewidth]{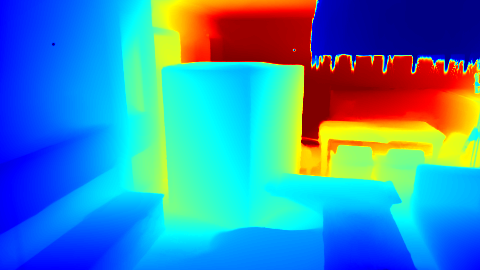}
    \end{subfigure}
    \begin{subfigure}{0.32\linewidth}
        \includegraphics[width=\linewidth]{fig_sources/supl/qual_ablation_geo/wonormal_scene_4_normal_0016.png}
    \end{subfigure}

    (c) w/o $\mathcal{L}_{\text{normal}}$
    \vspace{2pt}

    \begin{subfigure}{0.32\linewidth}
        \includegraphics[width=\linewidth]{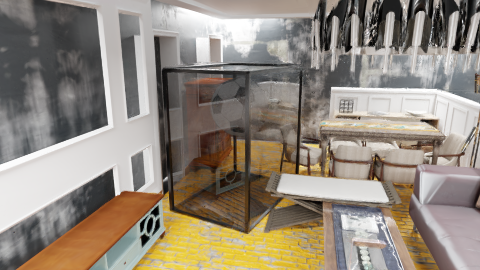}
    \end{subfigure}
    \begin{subfigure}{0.32\linewidth}
        \includegraphics[width=\linewidth]{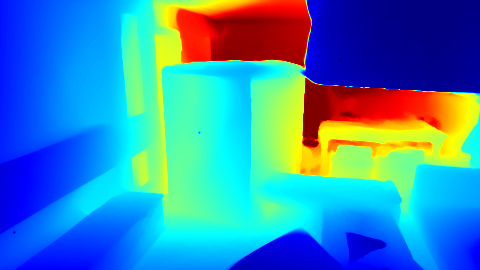}
    \end{subfigure}
    \begin{subfigure}{0.32\linewidth}
        \includegraphics[width=\linewidth]{fig_sources/supl/qual_ablation_geo/wogeo_scene_4_normal_0016.png}
    \end{subfigure}

    (d) w/o $\mathcal{L}_{\text{geo}}$

    \caption{\textbf{Effect of geometric losses.}
    Ablating geometric supervision leads to degraded geometry reconstruction.
    Removing $\mathcal{L}_{\text{depth}}$ produces inaccurate interface depth, 
    removing $\mathcal{L}_{\text{normal}}$ results in unstable surface orientation,
    and removing all geometric losses ($\mathcal{L}_{\text{geo}}$) severely distorts both depth and normals.}
    \label{fig:supl_qual_ablation_geo}
\end{figure*}

%% file: figures/supl/supl_qual_trans.tex
\begin{figure}[t!]
    \centering
    \setlength{\tabcolsep}{2pt}

    \begin{tabular}{ccc}
    \end{tabular}

    \vspace{4pt}

    \begin{subfigure}{0.32\linewidth}
        \includegraphics[width=\linewidth]{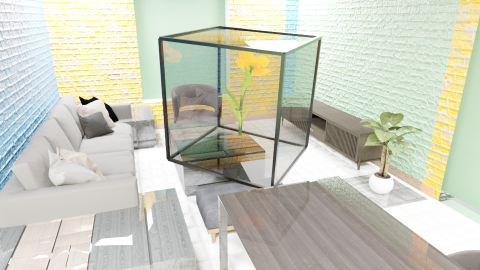}
    \end{subfigure}
    \begin{subfigure}{0.32\linewidth}
        \includegraphics[width=\linewidth]{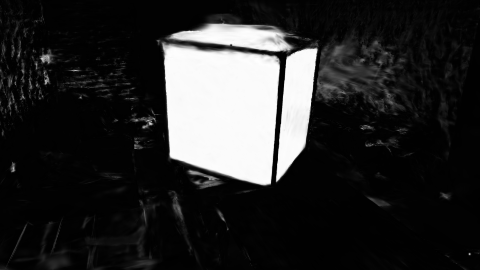}
    \end{subfigure}
    \begin{subfigure}{0.32\linewidth}
        \includegraphics[width=\linewidth]{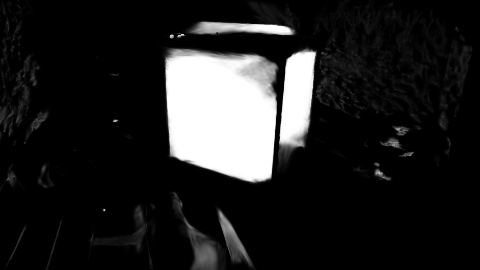}
    \end{subfigure}

    \begin{subfigure}{0.32\linewidth}
        \includegraphics[width=\linewidth]{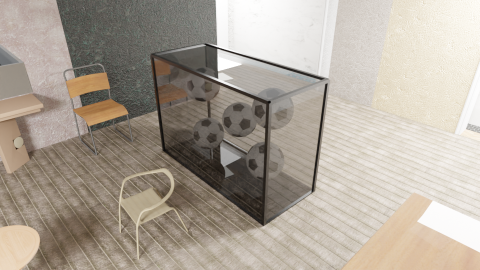}
    \end{subfigure}
    \begin{subfigure}{0.32\linewidth}
        \includegraphics[width=\linewidth]{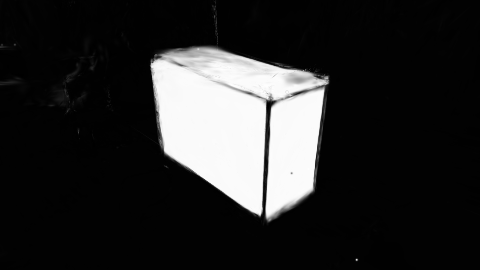}
    \end{subfigure}
    \begin{subfigure}{0.32\linewidth}
        \includegraphics[width=\linewidth]{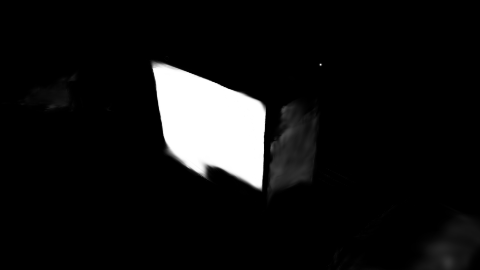}
    \end{subfigure}

    \caption{\textbf{Effect of the transparency loss $\mathcal{L}_{\text{trans}}$.}
    With the proposed transparency loss $\mathcal{L}_{\text{trans}}$, 
    the learned transparency maps align well with the true transmitted content,
    while removing this loss leads to noisy or inconsistent transparency estimation.}
    \label{fig:trans_ablation}
\end{figure}

%% file: figures/supl/radiance_decomp_compare.tex
\begin{figure*}[t!]
\centering
\setlength{\tabcolsep}{2pt}

\begin{subfigure}[t]{0.48\linewidth}
    \centering
    \begin{tabular}{cc}
        \includegraphics[width=0.48\linewidth]{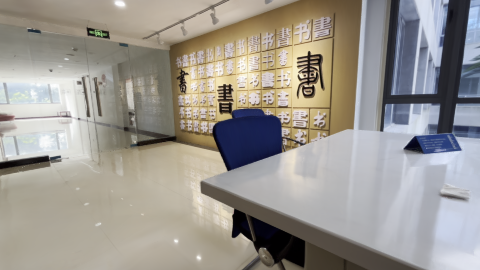} &
        \includegraphics[width=0.48\linewidth]{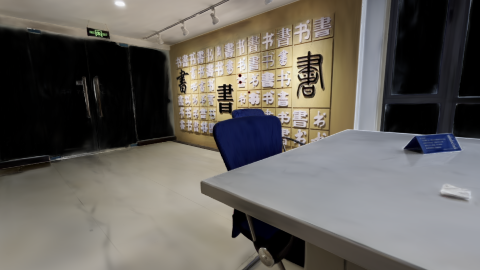} \\
        \includegraphics[width=0.48\linewidth]{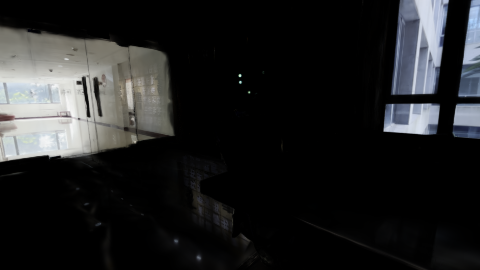} &
        \includegraphics[width=0.48\linewidth]{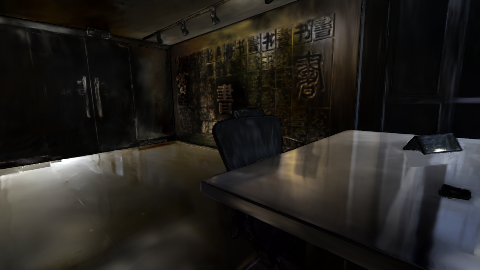}
    \end{tabular}
    \\[4pt]
    (a) Ours
\end{subfigure}
\hfill
\begin{subfigure}[t]{0.48\linewidth}
    \centering
    \begin{tabular}{cc}
        \includegraphics[width=0.48\linewidth]{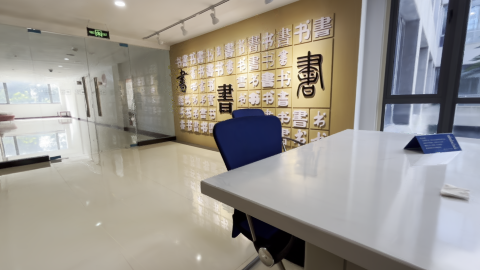} &
        \includegraphics[width=0.48\linewidth]{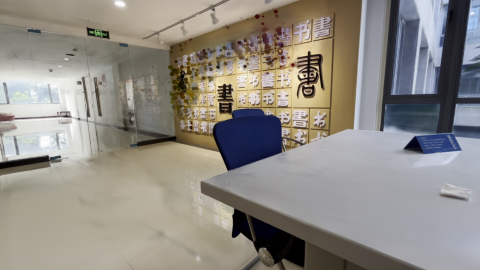} \\
        \includegraphics[width=0.48\linewidth]{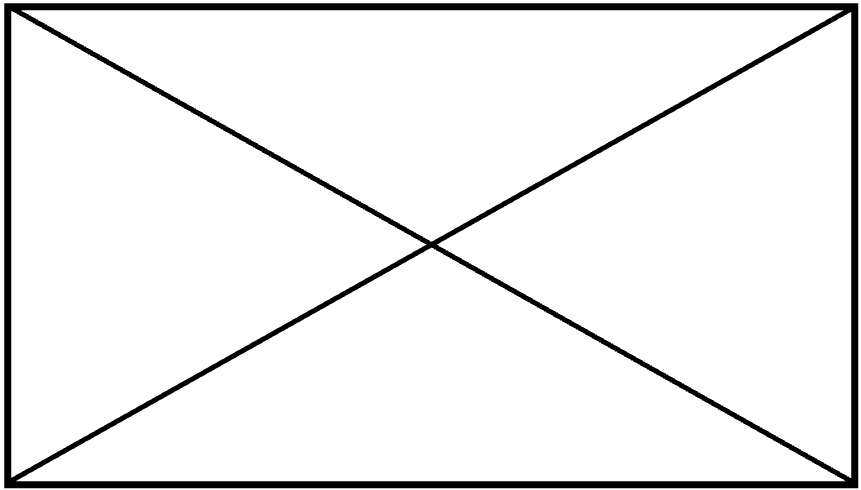}
        & \includegraphics[width=0.48\linewidth]{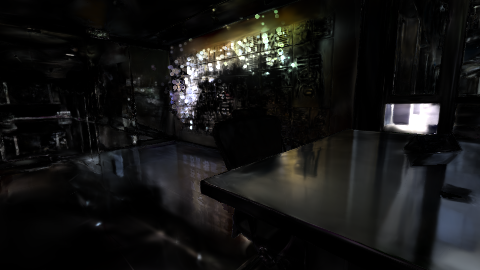}
    \end{tabular}
    \\[4pt]
    (b) EnvGS~\cite{xie2025envgs}
\end{subfigure}

\caption{
\textbf{Radiance decomposition comparison.}
Each $2 \times 2$ grid is arranged in clockwise order as rendered RGB, base color, reflection and transmission. EnvGS does not provide a transmission component, so the corresponding slot is left blank.  
}
\label{fig:radiance_decomp}
\end{figure*}

%% file: figures/supl/envgswdiffren.tex
\begin{figure}[t]
    \centering
    \setlength{\tabcolsep}{2pt}

    \begin{subfigure}{0.32\linewidth}
        \centering
        \includegraphics[width=\linewidth]{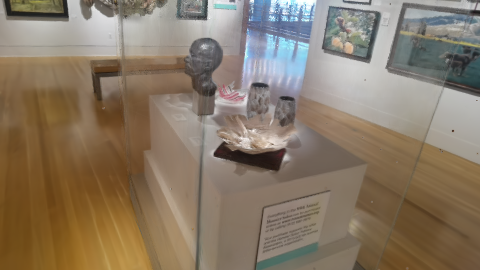}
    \end{subfigure}
    \begin{subfigure}{0.32\linewidth}
        \centering
        \includegraphics[width=\linewidth]{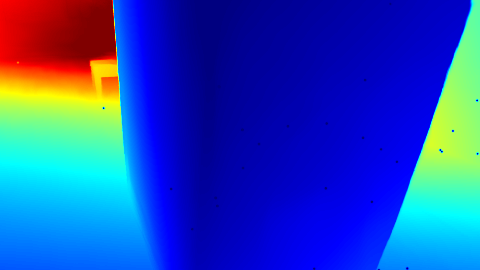}
    \end{subfigure}
    \begin{subfigure}{0.32\linewidth}
        \centering
        \includegraphics[width=\linewidth]{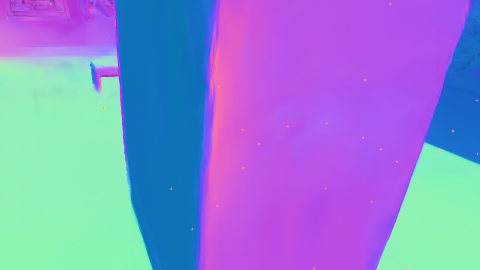}
    \end{subfigure}

    \vspace{2pt}
    \small (a) Ours
    \vspace{6pt}

    \begin{subfigure}{0.32\linewidth}
        \centering
        \includegraphics[width=\linewidth]{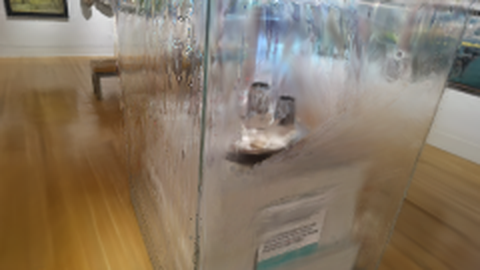}
    \end{subfigure}
    \begin{subfigure}{0.32\linewidth}
        \centering
        \includegraphics[width=\linewidth]{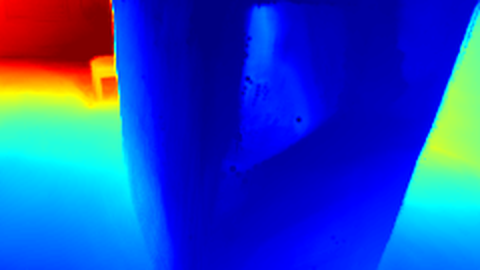}
    \end{subfigure}
    \begin{subfigure}{0.32\linewidth}
        \centering
        \includegraphics[width=\linewidth]{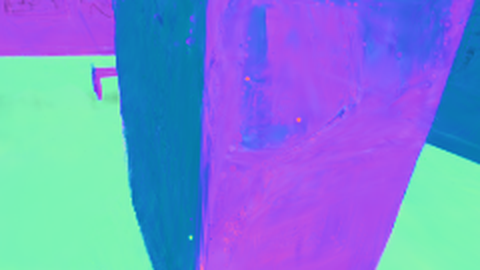}
    \end{subfigure}

    \vspace{2pt}
    \small (b) EnvGS~\cite{xie2025envgs} w/ DiffusionRenderer~\cite{liang2025diffusion}

    \caption{\textbf{Comparison with EnvGS~\cite{xie2025envgs} trained with a DiffusionRenderer~\cite{liang2025diffusion} prior.}
    With the DiffusionRenderer normal priors, EnvGS still fails to produce consistent depth and normals for transparent regions. In particular, the transmission surface remains incorrectly reconstructed, showing blurry rendering quality.}
    \label{fig:envgs_diffren_compare}
\end{figure}

%% file: figures/supl/supl_sam_comparison.tex
\begin{figure*}[t]
    \centering
    \setlength{\tabcolsep}{2pt}

    \begin{tabular}{c@{}c}

    \begin{tabular}{ccc}
        Reference & Ours & G-SAM2~\cite{ren2024grounded} \\[3pt]

        \includegraphics[width=0.16\textwidth]{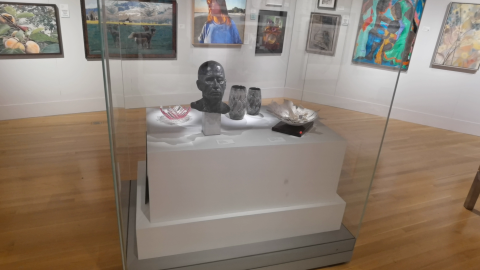} &
        \includegraphics[width=0.16\textwidth]{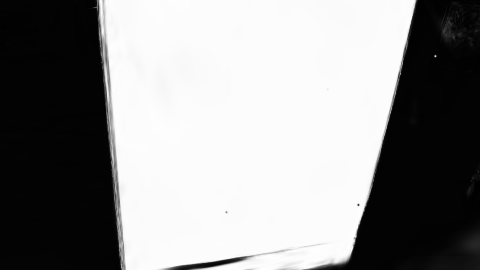} &
        \includegraphics[width=0.16\textwidth]{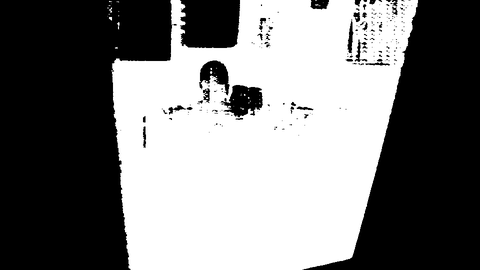} \\[-1pt]

        \includegraphics[width=0.16\textwidth]{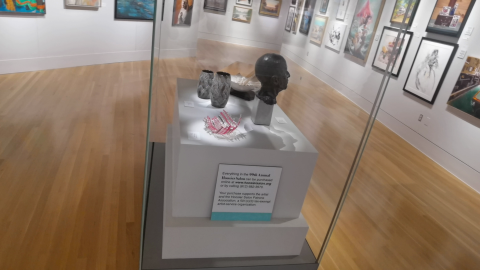} &
        \includegraphics[width=0.16\textwidth]{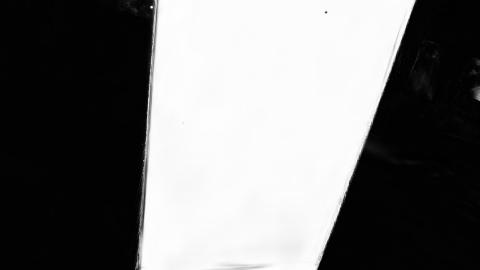} &
        \includegraphics[width=0.16\textwidth]{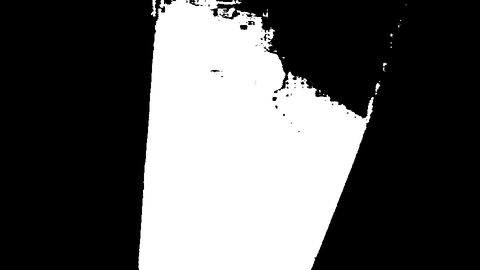} \\[-1pt]

        \includegraphics[width=0.16\textwidth]{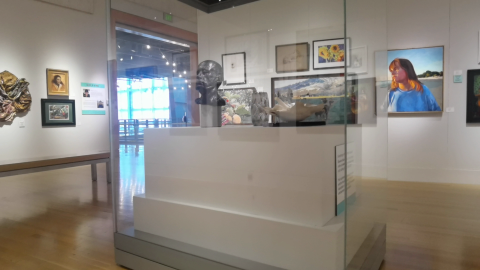} &
        \includegraphics[width=0.16\textwidth]{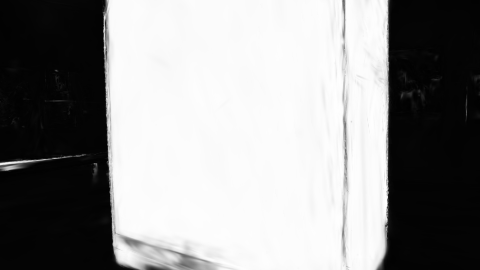} &
        \includegraphics[width=0.16\textwidth]{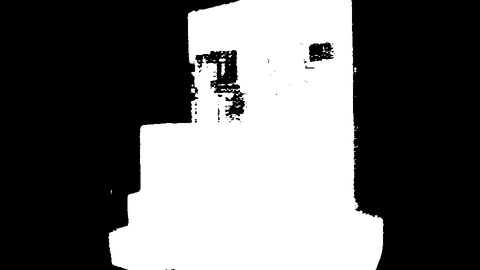}
    \end{tabular}
    &
    \begin{tabular}{ccc}
        Reference & Ours & G-SAM2~\cite{ren2024grounded} \\[3pt]

        \includegraphics[width=0.16\textwidth]{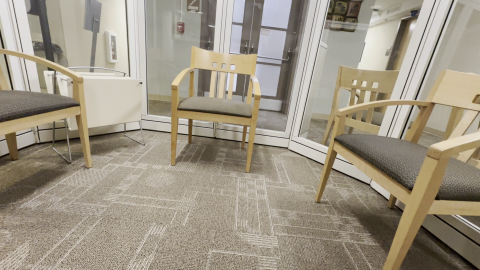} &
        \includegraphics[width=0.16\textwidth]{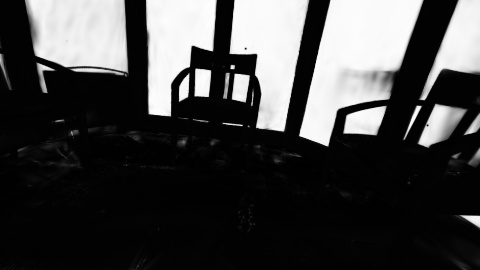} &
        \includegraphics[width=0.16\textwidth]{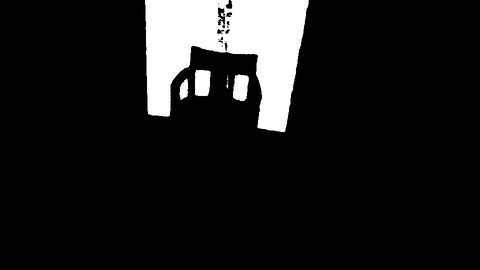} \\[-1pt]

        \includegraphics[width=0.16\textwidth]{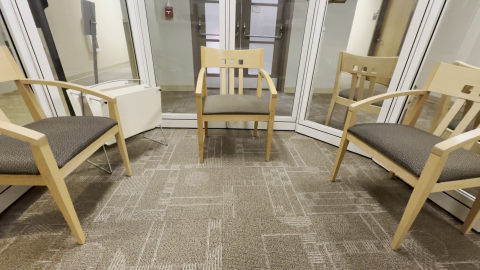} &
        \includegraphics[width=0.16\textwidth]{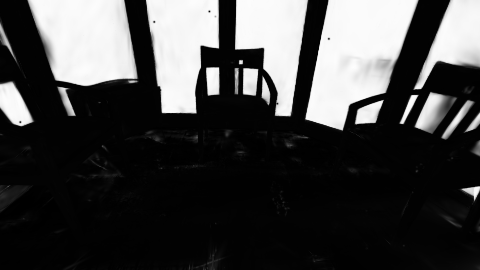} &
        \includegraphics[width=0.16\textwidth]{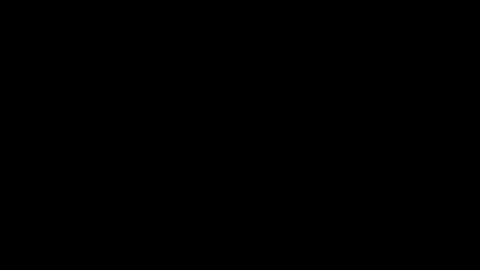} \\[-1pt]

        \includegraphics[width=0.16\textwidth]{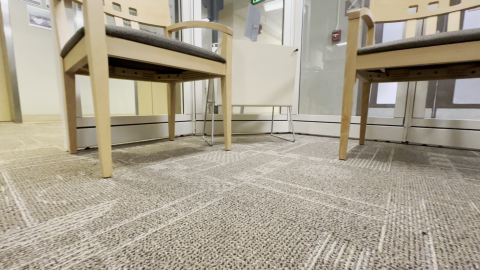} &
        \includegraphics[width=0.16\textwidth]{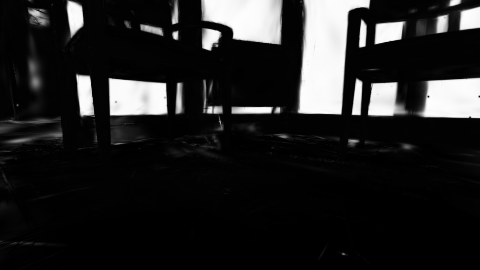} &
        \includegraphics[width=0.16\textwidth]{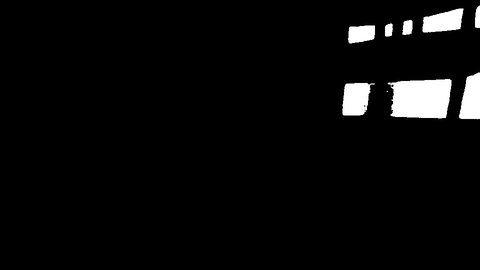}
    \end{tabular}

    \end{tabular}

    \caption{Comparison of transparency masks across two scenes.
    Each scene is shown with three viewpoints (rows), including ground-truth reference RGB (left),
    our predicted transparency maps (middle), and G-SAM2~\cite{ren2024grounded} masks (right).
    GT images provide visual context, highlighting that our method consistently isolates transparent surfaces,
    while Grounded-SAM2 often produces noisy, incomplete or completely missing masks.}
    \label{fig:trans_mask_two_scenes}
\end{figure*}

%% file: figures/supl/supl_qual_3d_front_t.tex
\begin{figure*}[t!]
\centering
\setlength{\tabcolsep}{1pt}

\begin{center}
\renewcommand{\arraystretch}{0.9}
\begin{tabular}{ccccc}
\makebox[0.19\linewidth][c]{{\textrm{GT}}} &
\makebox[0.19\linewidth][c]{{\textrm{PGSR}}~\cite{chen2024pgsr}} &
\makebox[0.19\linewidth][c]{{\textrm{EnvGS}}~\cite{xie2025envgs}} &
\makebox[0.19\linewidth][c]{{\textrm{TSGS}}~\cite{li2025tsgs}} &
\makebox[0.19\linewidth][c]{{\textrm{Ours}}} \\
\end{tabular}
\end{center}

\vspace{-7pt}

\begin{subfigure}[t]{\linewidth}
\centering
\includegraphics[width=0.195\linewidth]{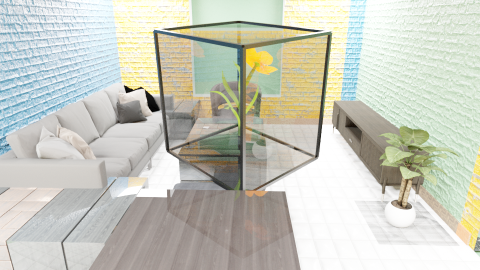}\hspace{1pt}%
\includegraphics[width=0.195\linewidth]{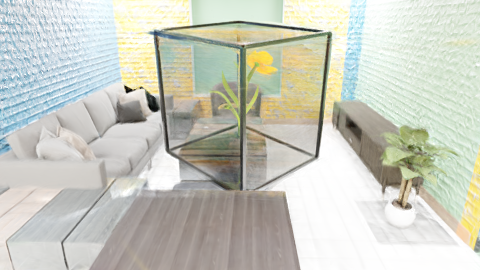}\hspace{1pt}%
\includegraphics[width=0.195\linewidth]{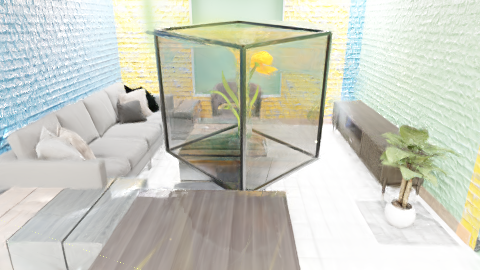}\hspace{1pt}%
\includegraphics[width=0.195\linewidth]{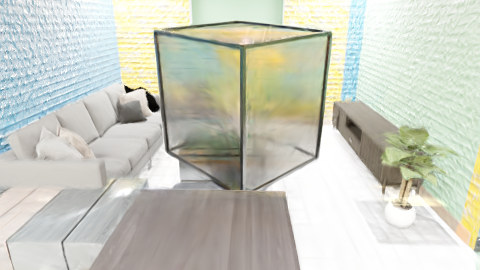}\hspace{1pt}%
\includegraphics[width=0.195\linewidth]{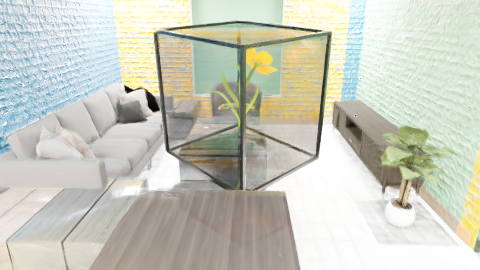}\\[1pt]

\includegraphics[width=0.195\linewidth]{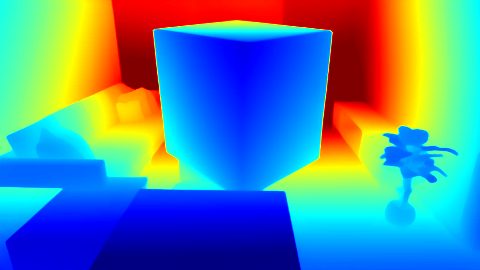}\hspace{1pt}%
\includegraphics[width=0.195\linewidth]{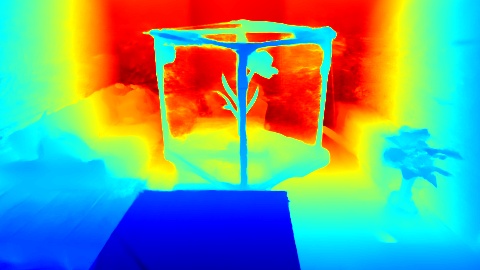}\hspace{1pt}%
\includegraphics[width=0.195\linewidth]{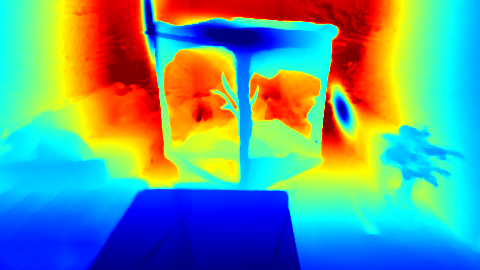}\hspace{1pt}%
\includegraphics[width=0.195\linewidth]{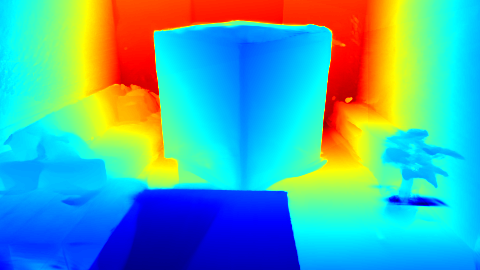}\hspace{1pt}%
\includegraphics[width=0.195\linewidth]{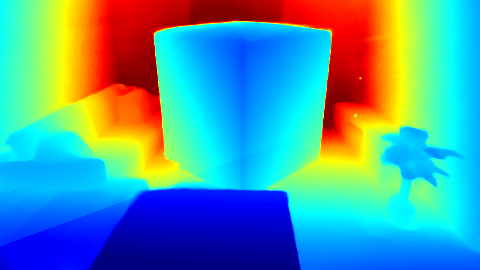}\\[1pt]

\includegraphics[width=0.195\linewidth]{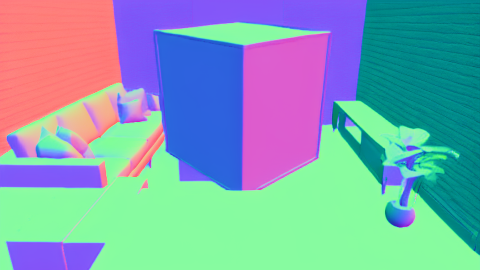}\hspace{1pt}%
\includegraphics[width=0.195\linewidth]{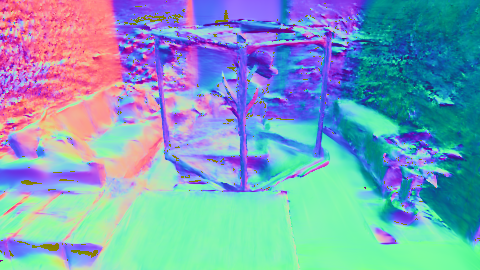}\hspace{1pt}%
\includegraphics[width=0.195\linewidth]{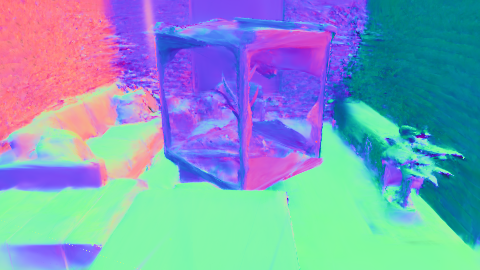}\hspace{1pt}%
\includegraphics[width=0.195\linewidth]{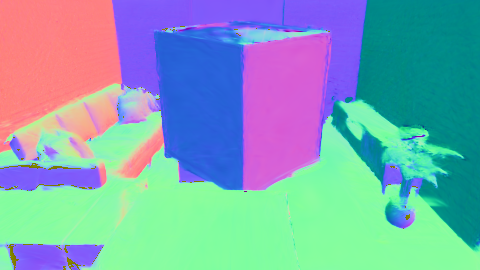}\hspace{1pt}%
\includegraphics[width=0.195\linewidth]{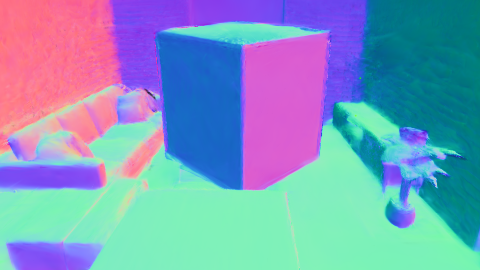}
\end{subfigure}

\vspace{2pt}\hrule\vspace{2pt}

\begin{subfigure}[t]{\linewidth}
\centering
\includegraphics[width=0.195\linewidth]{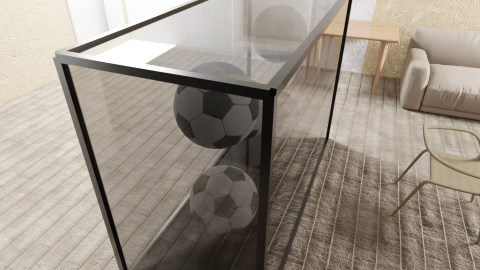}\hspace{1pt}%
\includegraphics[width=0.195\linewidth]{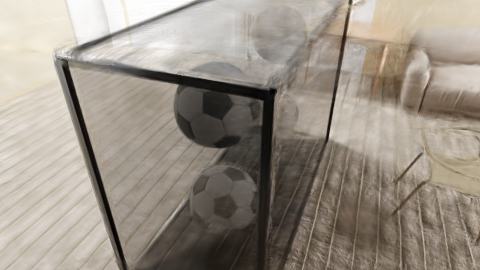}\hspace{1pt}%
\includegraphics[width=0.195\linewidth]{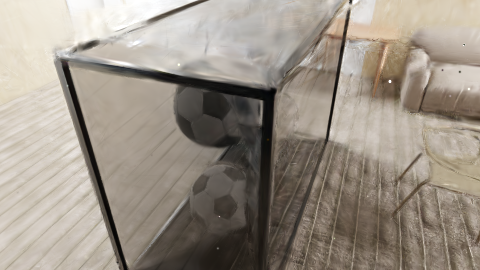}\hspace{1pt}%
\includegraphics[width=0.195\linewidth]{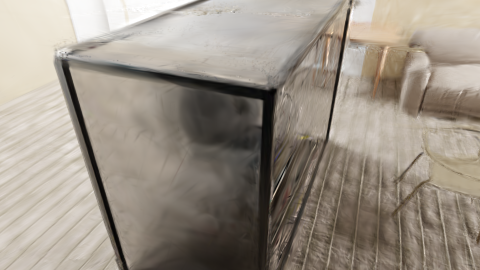}\hspace{1pt}%
\includegraphics[width=0.195\linewidth]{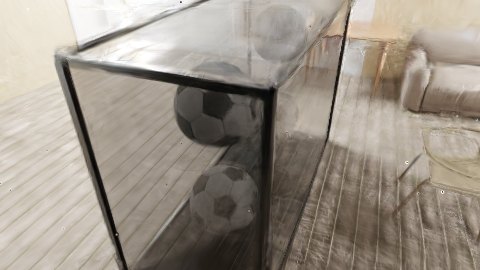}\\[1pt]

\includegraphics[width=0.195\linewidth]{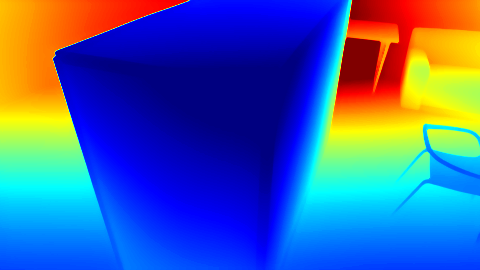}\hspace{1pt}%
\includegraphics[width=0.195\linewidth]{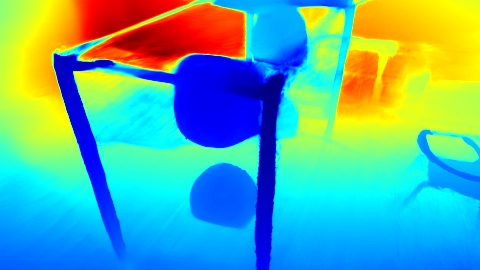}\hspace{1pt}%
\includegraphics[width=0.195\linewidth]{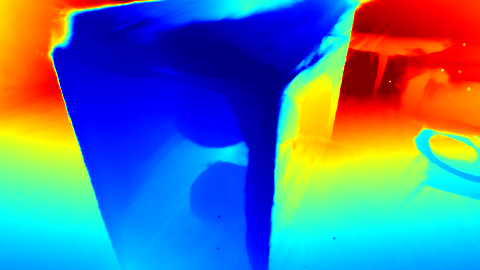}\hspace{1pt}%
\includegraphics[width=0.195\linewidth]{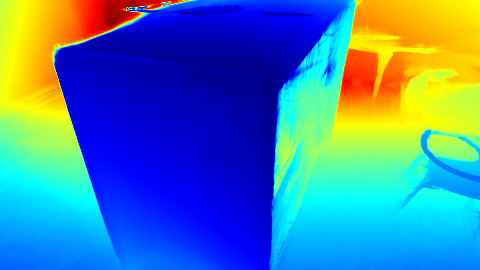}\hspace{1pt}%
\includegraphics[width=0.195\linewidth]{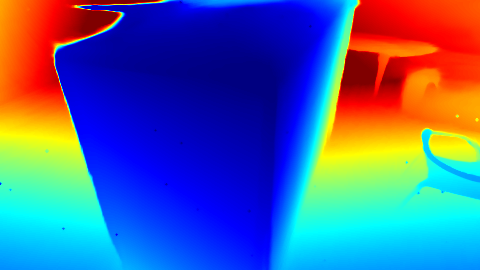}\\[1pt]

\includegraphics[width=0.195\linewidth]{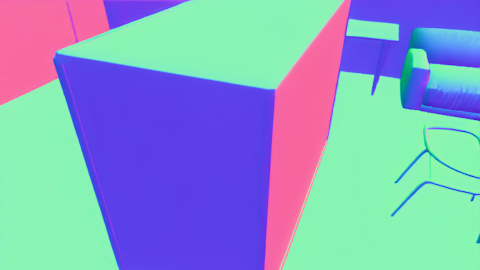}\hspace{1pt}%
\includegraphics[width=0.195\linewidth]{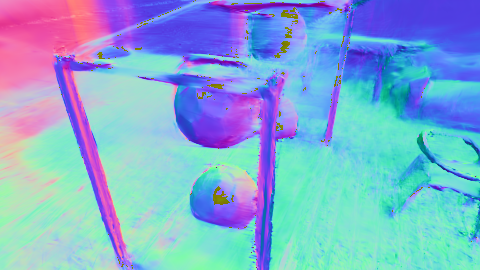}\hspace{1pt}%
\includegraphics[width=0.195\linewidth]{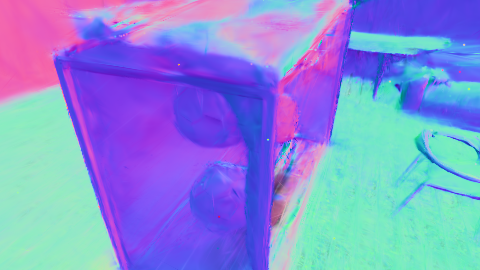}\hspace{1pt}%
\includegraphics[width=0.195\linewidth]{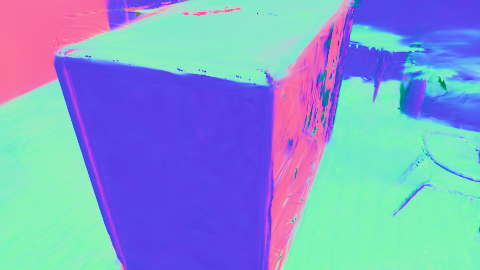}\hspace{1pt}%
\includegraphics[width=0.195\linewidth]{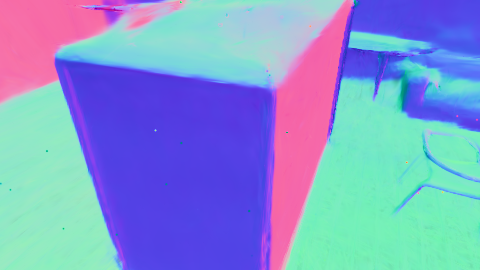}
\end{subfigure}

\vspace{-3pt}
\caption{
\textbf{Qualitative comparison on synthetic scenes.}
Each column shows results from GT, PGSR~\cite{chen2024pgsr}, EnvGS~\cite{xie2025envgs}, TSGS~\cite{li2025tsgs}, and Ours.
For each scene, rows correspond to RGB (top), depth (middle), and normal (bottom) maps.
}
\vspace{-7pt}
\label{fig:supl_additional_qual_3d_front_t}
\end{figure*}

%% file: figures/supl/supl_qual_dl3dv.tex
\begin{figure*}[t!]
\centering
\setlength{\tabcolsep}{1pt}

\renewcommand{\arraystretch}{0.9}
\begin{tabular}{ccccc}
\makebox[0.19\linewidth][c]{\textrm{GT}} &
\makebox[0.19\linewidth][c]{\textrm{PGSR}~\cite{chen2024pgsr}} &
\makebox[0.19\linewidth][c]{\textrm{EnvGS}~\cite{xie2025envgs}} &
\makebox[0.19\linewidth][c]{\textrm{TSGS}~\cite{li2025tsgs}} &
\makebox[0.19\linewidth][c]{\textrm{Ours}}
\end{tabular}


\begin{subfigure}[t]{\linewidth}
\centering

\includegraphics[width=0.195\linewidth]{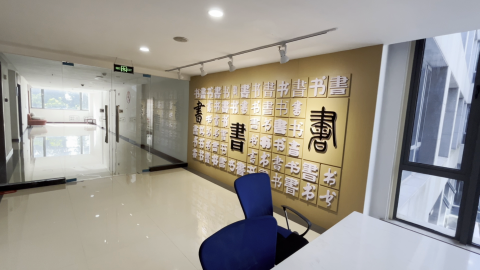}\hspace{1pt}%
\includegraphics[width=0.195\linewidth]{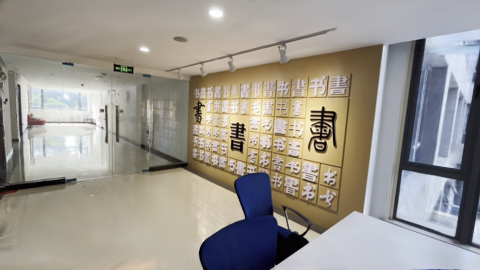}\hspace{1pt}%
\includegraphics[width=0.195\linewidth]{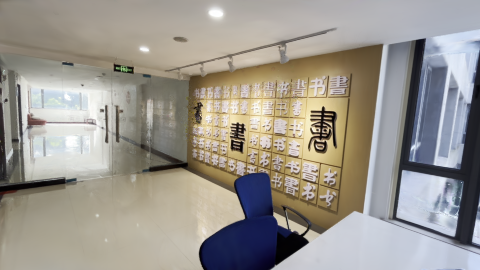}\hspace{1pt}%
\includegraphics[width=0.195\linewidth]{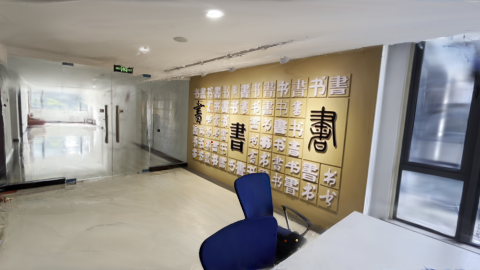}\hspace{1pt}%
\includegraphics[width=0.195\linewidth]{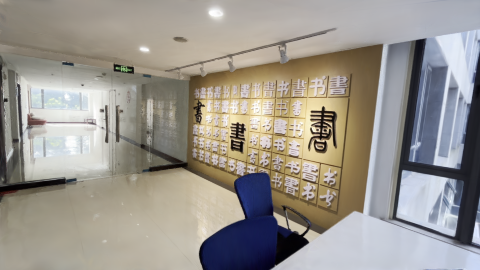}\\[2pt]

\makebox[0.195\linewidth][c]{}%
\includegraphics[width=0.195\linewidth]{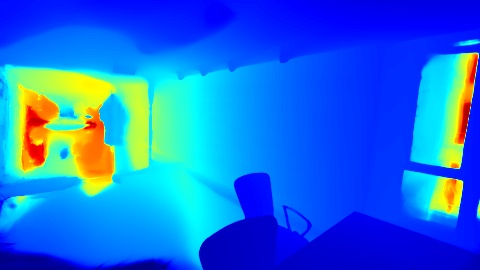}\hspace{1pt}%
\includegraphics[width=0.195\linewidth]{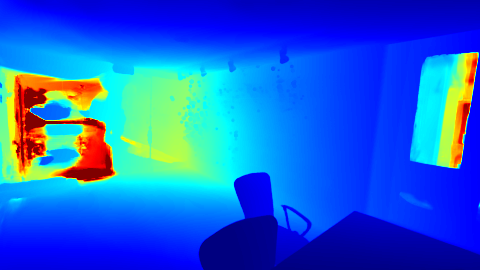}\hspace{1pt}%
\includegraphics[width=0.195\linewidth]{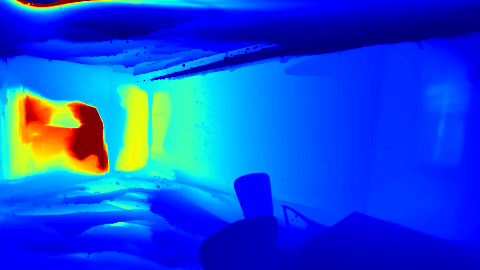}\hspace{1pt}%
\includegraphics[width=0.195\linewidth]{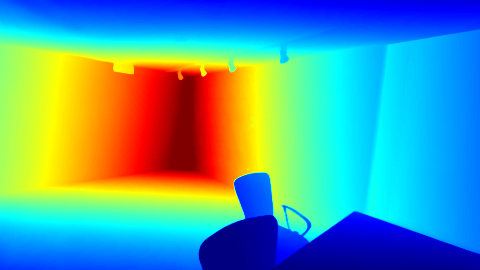}\\[2pt]

\makebox[0.195\linewidth][c]{}%
\includegraphics[width=0.195\linewidth]{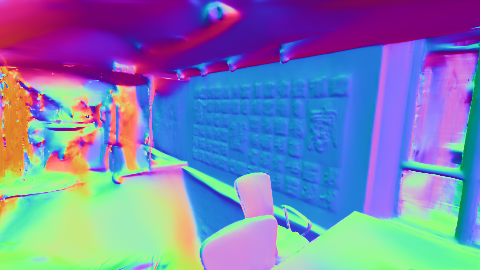}\hspace{1pt}%
\includegraphics[width=0.195\linewidth]{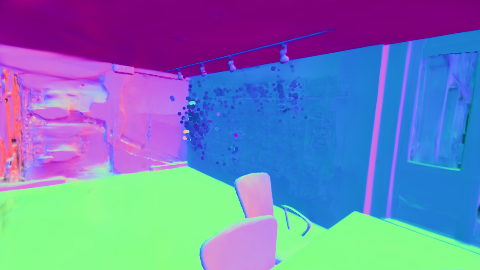}\hspace{1pt}%
\includegraphics[width=0.195\linewidth]{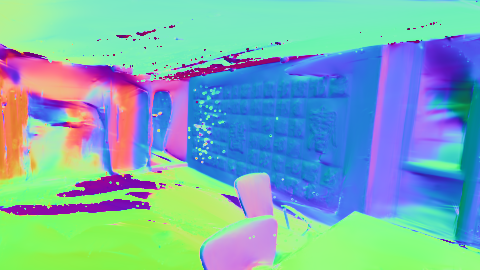}\hspace{1pt}%
\includegraphics[width=0.195\linewidth]{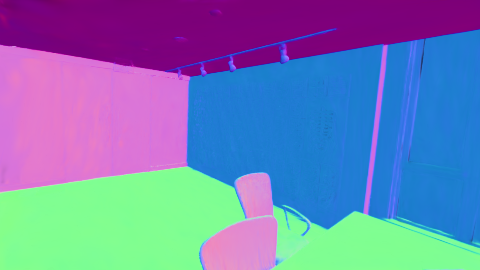}

\end{subfigure}

\vspace{2pt}
\hrule
\vspace{2pt}

\begin{subfigure}[t]{\linewidth}
\centering

\includegraphics[width=0.195\linewidth]{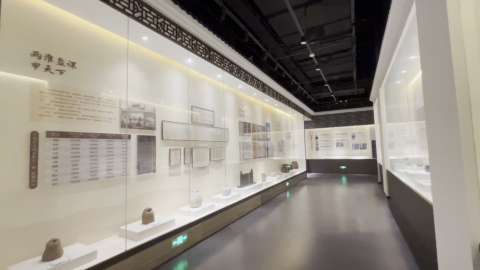}\hspace{1pt}%
\includegraphics[width=0.195\linewidth]{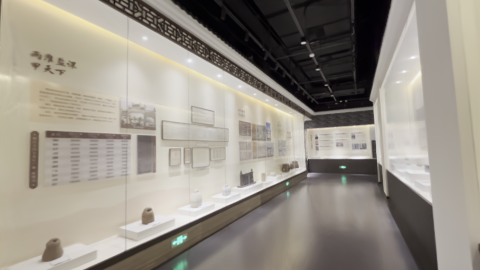}\hspace{1pt}%
\includegraphics[width=0.195\linewidth]{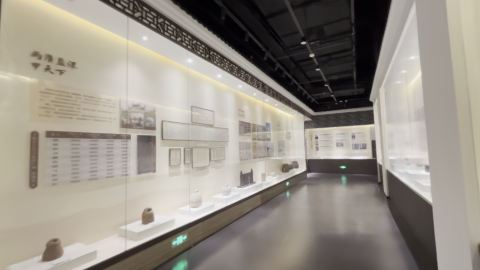}\hspace{1pt}%
\includegraphics[width=0.195\linewidth]{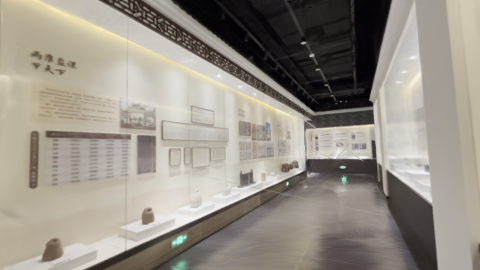}\hspace{1pt}%
\includegraphics[width=0.195\linewidth]{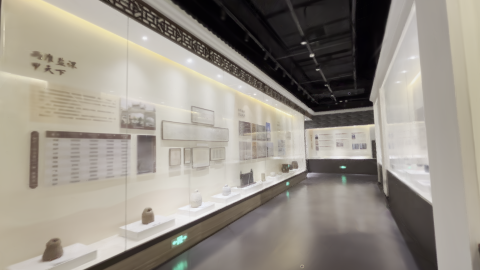}\\[2pt]

\makebox[0.195\linewidth][c]{}%
\includegraphics[width=0.195\linewidth]{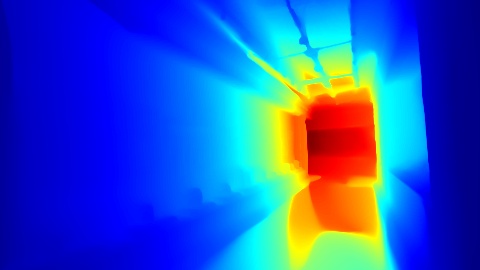}\hspace{1pt}%
\includegraphics[width=0.195\linewidth]{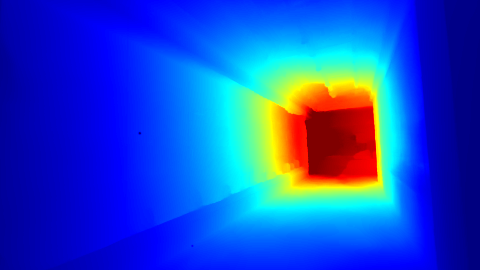}\hspace{1pt}%
\includegraphics[width=0.195\linewidth]{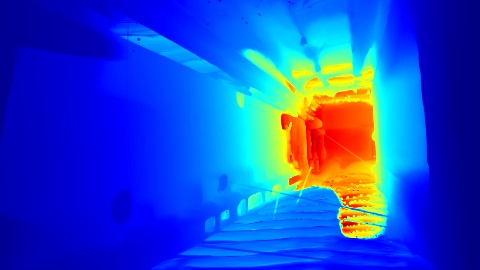}\hspace{1pt}%
\includegraphics[width=0.195\linewidth]{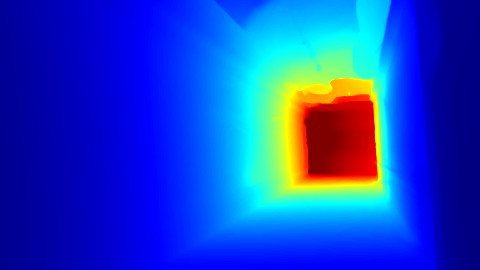}\\[2pt]

\makebox[0.195\linewidth][c]{}%
\includegraphics[width=0.195\linewidth]{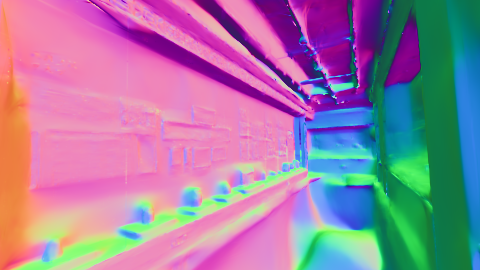}\hspace{1pt}%
\includegraphics[width=0.195\linewidth]{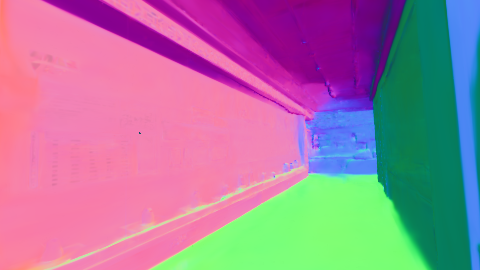}\hspace{1pt}%
\includegraphics[width=0.195\linewidth]{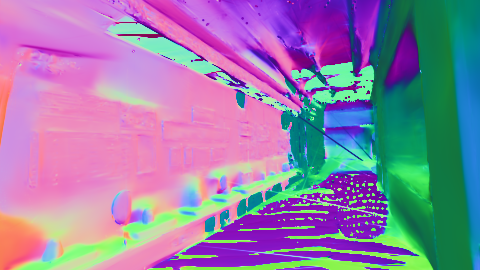}\hspace{1pt}%
\includegraphics[width=0.195\linewidth]{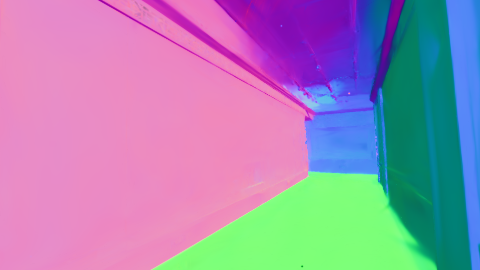}

\end{subfigure}

\vspace{-3pt}
\caption{
\textbf{Additional qualitative comparisons on the DL3DV-10K dataset.}
Each column shows results from GT, PGSR~\cite{chen2024pgsr}, EnvGS~\cite{xie2025envgs}, TSGS~\cite{li2025tsgs}, and Ours.
For each scene, rows correspond to RGB (top), depth (middle), and normal (bottom) predictions.
}
\label{fig:supl_additional_qual_dl3dv}
\vspace{-6pt}
\end{figure*}